\documentclass{draft}

\title{A Computational Model of the Institutional Analysis and Development Framework}
\titleheader{A Computational Model of the IAD Framework}
\author{Nieves Montes}
\authorabbrv{N. Montes}
\date{
	Artificial Intelligence Research Institute (IIIA-CSIC)\\
	Campus de la UAB\\
	Carrer de Can Planas, Zona 2\\
	08193 Cerdanyola del Vall\`{e}s, Barcelona\\
	nmontes@iiia.csic.es\\
	\today
	}

\usepackage{amsthm}
\theoremstyle{definition}
\newtheorem{definition}{Definition}

\newcommand\tab{\hspace*{0.5cm}}

\usepackage{todonotes}
\usepackage{url}
\usepackage{rotating}
\usepackage{tabularx}

\usepackage{adjustbox}

\begin{document}

\maketitle

\tableofcontents

\newpage

\section{Introduction}\label{sec:intro}
The Institutional Analysis and Development (IAD) framework is a conceptual toolbox put forward by Elinor Ostrom and colleagues in an effort to identify and delineate the universal common variables that structure the immense variety of human interactions \parencite{Ostrom2011}. The framework identifies {\em rules} as one of the core concepts to determine the structure of interactions, and acknowledges their potential to steer a community towards more beneficial and socially desirable outcomes.

This work presents the first attempt to turn the IAD framework into a computational model to allow communities of agents to formally perform {\em what-if} analysis on a given rule configuration. To do so, we define the Action Situation Language -- or ASL-- whose syntax is hgighly tailored to the components of the IAD framework and that we use to write descriptions of social interactions. ASL is complemented by a game engine that generates its semantics as an extensive-form game. These models, then, can be analyzed with the standard tools of game theory to predict which outcomes are being most incentivized, and evaluated according to their socially relevant properties.\footnote{All the code to go along with this work can be downloaded from \url{https://github.com/nmontesg/norms-games}.}

The remain of this report is organized as follows. We start by presenting the necessary background on the IAD framework, outline our contributions and review some related work in the rest of this Introduction. Then, we present the syntax of ASL in \Cref{sec:asl-syntax}. We explain in detail the process of rule interpretation in \Cref{sec:rule-interpretation}, a crucial step to turn action situation descriptions into games. Next, we go through the game semantics generation process in \Cref{sec:semantics}. The last technical part is \Cref{sec:implementation}, where we review some issues related to implementation and the integration of the resulting game representations with game-theoretical tools. Finally, we present some illustrative examples in \Cref{sec:examples} and conclude in \Cref{sec:conclusions}.

\subsection{The Institutional Analysis and Development framework}\label{subsec:iad}
\begin{figure}
	\centering
	\includegraphics[width=\linewidth]{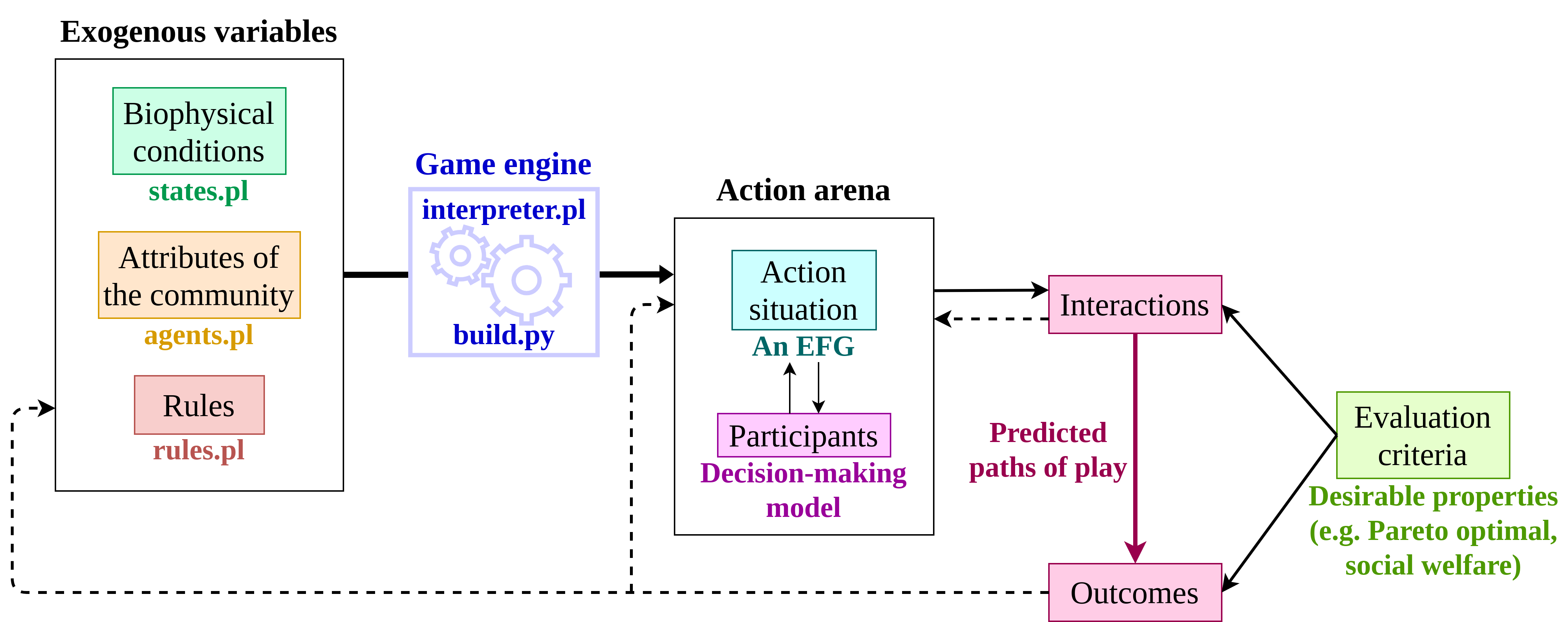}
	\caption{Outline of the Institutional Analysis and Development framework. Adapted from \cite[p. 15]{Ostrom2005}. Colored text outside boxes indicate either the scripts that contain information on the boxed component, or the game-theoretical concepts that represent it.}
	\label{fig:iadframework}
\end{figure}

Within the field of policy analysis, the Institutional Analysis and Development (IAD) framework, put forward by Ostrom and colleagues \cite{Ostrom2005}, represents a comprehensive theoretical effort to identify and delineate the universal building blocks that make up any social interaction. Its outline is presented in \Cref{fig:iadframework}. In the center part, any social interaction is referred to as an {\em action arena}. In it, a set of {\em participants} (the agents) find themselves in an {\em action situation}, which is the social space they may enter, take actions in and jointly bring about outcomes.

Action arenas are affected by three sets of exogenous variables that jointly combine to structure it. These are the {\em biophysical conditions}, the {\em attributes of the community} and the {\em rules} in use. The first two are fairly straightforward to define. The biophysical conditions refer to the relevant environmental characteristics of the space where the interaction takes place, such as land topology and location of resources. The attributes of the community include variables intrinsically linked to the participants, such as age, gender, ethnicity and/or belonging to one of multiple subgroups. Last of all, the term {\em rules} is wide enough to need a detailed clarification of the use we will make of it in this work.

There are four common uses of the term {\em rules} in everyday language \parencite[Ch. VI]{Black1962}: instructions, precepts, regulations and principles. First, instructions are understood as a set of steps to effectively achieve some desirable outcome. Good contemporary examples are Ikea assembly guides. In second place, precepts are somewhat similar to instructions, however their scope is more general. Instead of specifying the particular actions that an agent should perform, precepts provide abstract, widely applicable principles from which a course of action in a particular situations should be deduced. Precepts often appeal to moral values, such as ``you should act with compassion towards others''.

In the third place, regulations are, possibly, the most intuitive meaning of rules. They refer to statutes and ordinances that constrain or provide alternative avenues for a course of action. Typically, regulative rules are understood as being passed down from an authority responsible for their crafting and enforcement. However, small communities can also self-impose regulations on themselves in order to ensure sustainability, fairness, and other desirable goals. Finally, the last meaning that rules take are as principles. These refer to the laws of nature that inevitably play a part in determining what actions and/or outcomes are physically possible.

There is a major difference between the first two and the last two meanings. While instructions and precepts indicate to an agent (either directly or indirectly) what to do provided the situation at hand, regulations and principles condition the structure that the situation itself has. Together, regulations and principles jointly determine what actions are possible and/or allowed, what their effects are and, consequently, which outcomes may be attained. Instructions and precepts take in that information as an input, and output a particular course of action.

In this work, we take the view that the function of rules is to mold the structure of the situation agents find themselves in, i.e. to modify the incentives and opportunities they find themselves in. Hence, the term {\em rule} will be used to encapsulate both regulations and natural principles. There is a fundamental difference, however, between the two. While regulations are human-made, and hence subject to revision and change, natural principles are not and therefore essentially untouchable. We address this distinction in our logical language by distinguishing between {\em default} rules and additional regulations through a priority integer linked to every rule statement. Also, we wish to make the distinction between rules that reflect natural principles and the biophysical conditions clear early on. Physical laws control the {\em dynamics} of the environment (drop an object and it will land on the ground) while biophysical conditions refer to static elements (like land topology).

\begin{figure}
	\centering
	\includegraphics[width=0.8\linewidth]{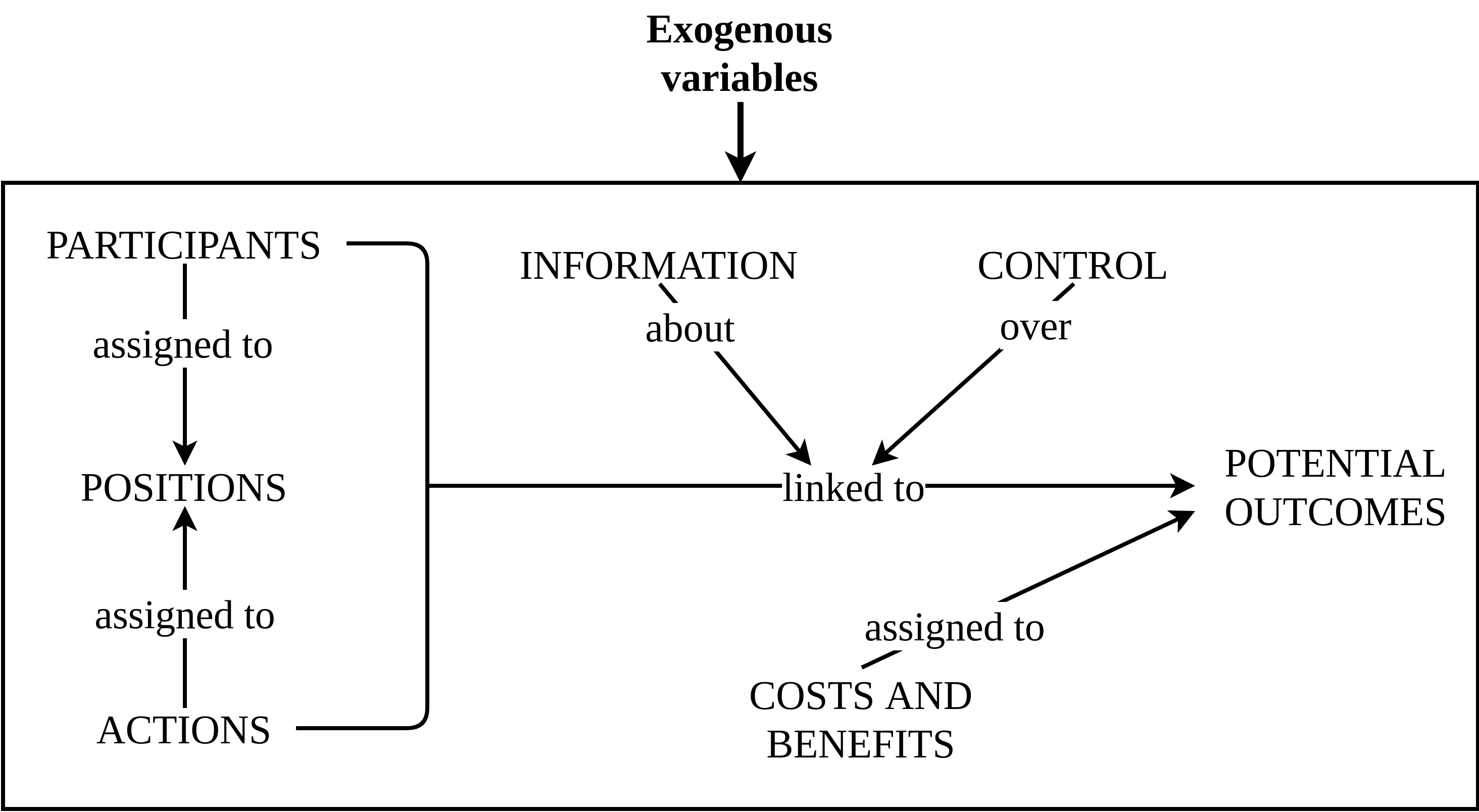}
	\caption{Internal structure of an action situation. Reproduced from \cite[p. 33]{Ostrom2005}.}
	\label{fig:action-situation}
\end{figure}

In addition to identifying the variables that condition an action arena, the IAD framework also determines the components that together make up any action situation. There are in total seven variables at play (see \Cref{fig:action-situation}): (1) the participants who are allowed to enter; (2) the positions or roles that they take on; (3) the actions assigned to those roles; (4) the potential outcomes that may be reached; (5) the linkage between actions (or sequences of actions) to outcomes and the control that agents have over it; (6) the information available to participants; and (7) the material reward and costs assigned to outcomes and/or actions.

Once participants populate an action situation, it becomes fully instantiated. By introducing some decision-making model for every agent (such as traditional rationality notions like the Nash equilibrium), a prediction of how the interaction is expected to play out and the eventual outcomes (the ``Interactions'' and ``Outcomes'' boxes in \Cref{fig:iadframework}) that are likely to be reached can be constructed. Finally, these outcomes can be evaluated in terms some desirable properties (``Evaluation criteria'' box), such as optimality, efficiency, or various metrics of social welfare \parencite{Weymark2016}.

If the evaluation is not satisfactory, changes to the exogenous variables should be made in pursue of a more successful outcome (according to the evaluation criteria of choice). Of the three sets of exogenous variables, biophysical conditions and the attributes of the community are viewed as fixed in the short term. However, rules are relatively malleable. In particular, human-crafted regulations are very much susceptible to review and modification. The ``default'' rules (those that reflect natural principles) cannot be changed.

Despite being introduced several decades ago \parencite{Kiser1982}, the IAD framework is currently being used in policy analysis studies. Most recent examples include scenarios of pollution and waste management in widely diverse areas of the world \parencite{Sarr2021, Nguyen2020} as well as conservation policy \parencite{Barton2017}. Research efforts on the theoretical front are also on-going in order to better integrate the formulation of the IAD framework with existing formal legal systems \parencite{Cole2017}.

\subsection{Contributions}\label{subsec:contributions}
The main contribution of this work is a computational model of Ostrom's IAD framework, to enable communities of agents to formally perform {\em what-if} analysis of potential new regulatory rules to adopt. We provide new tools, as well as borrow existing concepts, that together compose the complete pipeline from a rule specification to an evaluation of its impact with whichever metric of choice.

In order to write rule configurations in a systematic manner, we present our novel Action Situation Language (ASL). This is a machine-readable logical language (implemented in Prolog) whose syntax is highly tailored to the exogenous variables outlines in the IAD framework. ASL is complemented by a game engine that takes as input a valid action situation description and automatically generates its semantics as an extensive-form game (EFG). Although environmental and community attributes also play a role in generating the semantics, we are particularly interested in the impact that rules have on the resulting formal model. In fact, an essential component of the game engine is a rule interpreter, whose function is to query the rule base, process their implications and solve conflicts between contradicting rules.

In fact, the two main innovations we present (ASL plus its game engine) bridge the gap between the normative multiagent systems (norMAS) and game theory fields. In norMAS, a great deal of work has been devoted to the study of {\em norms}, {\em rules} and other constraining mechanisms to achieve coordination among autonomous agents \parencite{Shoham1995,Onn1997,Andrighetto2012}. In parallel, game theory has provided a powerful toolbox to model multiagent interactions of competitive, cooperative and hybrid nature. Very well established game theoretical solution concepts are prevalent across the MAS literature (e.g. \cite{Hahn2020,Caillou2009}). However, in game theory, the rules that configure the structure of the interaction become irrelevant once the formal model has been built, and are often expressed in non-systematic, plain natural language. With ASL, such rules can be expressed in a systematic manner and automatically linked to its semantically equivalent formal game by feeding them to the game engine.

The choice of extensive-form games as the semantics for an ASL description is motivated by the availability of many game-theoretical solution concepts, such as traditional rationality notions (e.g. Nash or correlated equilibrium, subgame perfect equilibria, etc) and social properties of outcomes (e.g. Pareto efficiency, social welfare), that can be readily applied to any model built by the game engine. Due to the prevalence of these well-established concepts in the game theory literature, we do not see the need to provide new solution concepts of our own. Introducing such models of agent decision-making (either ``rational'' in the traditional sense or not) amounts to modeling the {\em participant} component of an action arena (see \Cref{fig:iadframework}). This step then paves the way to compute the most likely outcomes and evaluating them according to their optimality, efficiency, or social welfare. At this point, the pipeline from the rule configuration to an evaluation of its impact is complete, and the community of agents involved is informed about the impact that such regulations would have on them, were they to be adopted.

\subsection{Related work}\label{subsec:related-work}
Originally, the IAD framework was complemented by the Institutional Grammar (IG) \parencite{Crawford1995}. The IG parses institutional statements (which include strategies, norms in the sense of conventions and regulative rules) into five fields: the attributes (A) of the participants to whom the statements applies; the deontic (D) modality (permitted, forbidden or obliged); the aim (I) of the statement, meaning the action or outcome to whom the deontic applies; the condition (C) under which the statement applies; and the or-else (O) field which states the consequences of not following a rule. Put together, these fields constitute the ADICO syntax. The three types of institutional statements are distinguished by the fields that are necessary to describe them: AIC for strategies, ADIC for conventions and ADICO for rules. Lately, the IG has spurred renewed interest, with extensions to the original proposal including the nesting of statements \parencite{Frantz2013} and the distinction between different levels of granularity in the parsing \parencite{Frantz2021}. 

Although the early version of the IG did contain examples of formal games built from institutional statements (see e.g. \cite[Ch. 5-6]{Ostrom2005}), no attention is paid at the automation this process, as the ADICO syntax is not designed as machine-readable language. However, one of its interesting features, which we will import into ASL to some extent, is the classification of rules based on the component of the action situation that they target, according to the aim (I) field.

Although not being machine-readable, some works do attempt to make the ADICO syntax operational in agent-based models (e.g. \cite{Ghorbani2016, Smajgl2008}). There, the ADICO syntax is used to represent agents' strategies and shared conventions. However, these works are limited in scope, since they use very restricted forms of institutional statements and only target the modeling of common-pool resource situations. Although the IAD framework indeed accounts for the analysis of this type of scenarios, it is intended to identify and analyze the components of any social interaction. Due to these limitations, in this work we move away from the ADICO syntax and define our own machine-readable language, which is able to model a wide variety of action situations by leveraging the generality of game theoretical models.

On another front, the field of General Game Playing (GGP) within the AI community has come up through the years with machine-processable languages for the specification of general games. Most prominently, the Game Description Language (GDL) \parencite{Genesereth2005} is a high-level language for the specification of games with a finite number of players and legal moves. GDL provides compact descriptions of deterministic classical games (such as chess and checkers) and also admits a form of restricted imperfect information in the form of simultaneous moves, as feature that we incorporate into our language.

Since its creation, some extensions have been added to GDL in order to improve its expressive power. Most notably, GDL-II \parencite{Schiffel2014} incorporated the possibility of imperfect information and random moves by nature, although limiting those to a uniform probability distribution over them. Later, yet another addition resulted in the introduction of GDL-III, where epistemic games in which the rules depend on the knowledge of the agents can be represented by introducing player introspection \parencite{Thielscher2016}. Beyond game playing, GDL (in its original version) has been used for more socially relevant applications, such as mediated dispute resolution \parencite{Jonge2017} and the definition of domains for automated negotiations \parencite{Jonge2021}.

Although both GDL and our ASL are logical language for game specification, some of the feature of ASL make it much more well-suited than GDL for modeling socioeconomic interactions. While the rules of the game are implicit in GDL descriptions, In ASL they are explicitly and represented one by one. Consequently, ASL descriptions are more declarative than GDL ones, as new rules can be easily added and their individual impact examined. As a result, ASL descriptions are less rigid than GDL ones, as they are intended to be expanded with new regulations that, most probably, generate a different game structure. Meanwhile, GDL descriptions are not really meant to be dynamic, nor modified in any way during run-time.

\section{ASL syntax}\label{sec:asl-syntax}
Our intention is to define the syntax of ASL to be fully machine-readable, yet also relatively syntactically friendly to make it accessible to social science scholars and professionals. In order to completely describe an action situation, our language must specify the three sets of exogenous variables that determine it (see \Cref{fig:iadframework}):
\begin{itemize}
	\item \textbf{Attributes of the community}: the agents susceptible of taking part in the interaction, plus any of their relevant characteristics they may have: age, gender, ethnicity, etc.
	\item \textbf{Biophysical and environmental conditions}: land topology, location of resources, etc.
	\item The \textbf{rules} structuring the situation, in particular the following four {\em types}:
	\begin{itemize}
		\item \textbf{Boundary rules}: what agents are allowed to enter the action situation. For example, in many countries it is required to be over 18 years old to participate in an electoral process.
		\item \textbf{Position rules}: what roles do the participants take on, like candidate, voter, etc.
		\item \textbf{Choice rules}: what actions are available to the various roles under the current conditions. For example, an agent with the role {\em voter} can take the action to vote for one (or none) of the candidates.
		\item \textbf{Control rules}: what are the effects of those actions. In a majority rule electoral process, the candidate with the most votes gets appointed to the position in contention.
	\end{itemize}
\end{itemize}

Additionally, the following information is also necessary:
\begin{itemize}
	\item The initial conditions when the interaction starts.
	\item The termination conditions under which the interaction halts.
	\item What facts describing the state of the action situation are compatible with one another and can be simultaneously true (for example, an agent cannot be at two different locations at the same time).
\end{itemize}

\begin{table}[t]
	\caption{Action Situation Language keywords, sorted into reserved predicate symbols (with their arity) and operators (with their type in parenthesis).}
	\label{tab:asl-keywords}
	\centering
	\begin{tabular}{|c|c|}
		\hline \textbf{Predicates} & \textbf{Operators} \\
		\hline 
		\begin{tabular}{ll}
			\texttt{agent/1}& \texttt{rule/4}\\
			\texttt{participates/1} & \texttt{role/2}\\
			\texttt{can/2} & \texttt{does/2}\\
			\texttt{initially/1} & \texttt{terminal/0}\\
			\texttt{incompatible/2} & \\
		\end{tabular} &
		\begin{tabular}{ll}
			\texttt{if} (prefix) & \texttt{then} (infix)\\
			\texttt{where} (infix) & $\sim$ (prefix)\\
			\texttt{withProb} (infix) & \texttt{and} (infix)
		\end{tabular}\\
		\hline
	\end{tabular}
\end{table}

The keywords of ASL are gathered in \Cref{tab:asl-keywords}. Most of these appear as part of \texttt{rule} arguments, and only \texttt{agent}, \texttt{initially}, \texttt{terminal} and \texttt{incompatible} are used as standalone predicates.\footnote{In fact, the \texttt{agent} predicate appears both within \texttt{rule} statement and as a standalone predicate.} We start by reviewing the predicates that do not appear within rules. First, \texttt{agent(A)} denotes \texttt{A} as an individual susceptible of entering the action situation. Thus this predicate provides information on the community attributes. If needed, domain-dependent predicates of the type \texttt{feature\_name(Agent,Value)} can be added to encode agent attributes (for example, \texttt{age(alice,34)}).

Second, \texttt{initially(F)} indicates that fact \texttt{F} holds true at the start of the interaction, prior to any action being executed. \texttt{terminal} plays the opposite role, as it returns true whenever the conditions for halting the interaction are met. Finally, \texttt{incompatible(F,L)} states that fact \texttt{F} cannot be simultaneously true with the fluents in list \texttt{L}.

\begin{figure}[t]
	\centering
	\begin{tabular}{rl}
		\texttt{Rule} ::= & rule( \\
		& \tab \texttt{Id},\\
		& \tab \texttt{Type},\\
		& \tab \texttt{Priority},\\
		& \tab if \texttt{Condition} then \texttt{Consequence} where \texttt{Constraints}\\
		&).\\
		\texttt{Type} ::= & boundary | position | choice | control \\
		\texttt{Priority} ::= & 0 | 1 | ... | $\infty$
	\end{tabular}
	\caption{General syntax for {\em if-then-where} rules.}
	\label{fig:general-rule-syntax}
\end{figure}

We move on now to \texttt{rule/4} predicates. All of its clauses, regardless of the component they target, follow the general template in \Cref{fig:general-rule-syntax}, with the following four arguments:
\begin{enumerate}
	\item An identifier \texttt{Id} that denotes the action situation where the rule is to be applied.
	\item The \texttt{Type} of the rule, one of either {\em boundary}, {\em position}, {\em choice} or {\em control}.
	\item The \texttt{Priority} of the rule. This is a non-negative integer that determines which statement is to be prevail in case several rules lead to contradicting effects. The rule statements that are supposed to reflect the physical principles of the domain are assigned priority 0 and are referred to as the {\em default rules}, while additional human-made regulations have strictly positive priorities. The reserved overwriting operator \verb|~| is introduced in order to have high priority rule nullify the effects of lower priority rules.\footnote{We use the term {\em overwriting} instead of {\em negation} operator since ASL, as a logic programming, follows the convention of negation as failure.}
	\item The content of the rule is expressed with an \texttt{if-then-where} statement (the three are all ASL reserved opretaors, see \Cref{tab:asl-keywords}). The content of the \texttt{Condition} and \texttt{Consequence} fields is subject to syntactic constraints according to the type of the rule in question. \texttt{Constraints} always consists of a list of literals whose free variables unify with those in \texttt{Condition} and \texttt{Consequence}. The separation of rule pre-conditions into a short \texttt{Condition} and a \texttt{Constraints} field is not technically indispensable, but rather a stylistic choice to help keep the syntax concise.
\end{enumerate}

Next, we review in detail the syntactic restrictions for every rule type.

\subsection{Syntax by rule type}\label{subsec:rules-syntax}
As introduced, ASL considers four rule types (boundary, position, choice and control) that target different action situation components (see \Cref{fig:action-situation}). First, the boundary rules are aimed at regulating the {\em participants} components of action situations, as they designate which agents is able to enter the interaction. Second, position rules are responsible for assigning participants, once they are in the action situation, to the {\em roles} they assume. A participants may take on multiple roles. However, a participants that is not assigned any role is irrelevant to the evolution of the interaction, as the {\em actions} that bring about the outcomes are assigned to roles through choice rules, and not to participants directly. Finally, control rules are responsible for the dynamics of the interaction.They map actions by the participants to their effects. Hence, this rule type is directly responsible for the {\em control} component of action situations.

Note that there is some disconnect between our four rules types and the seven variables within an action situation (\Cref{fig:action-situation}). There are not dedicated rule types for the outcomes, costs and benefits, and information variables. For the first two (outcomes and costs and benefits), we argue that control rules are in charge. As they effectively regulate how does the state of the world evolve, they are also indirectly determining what {\em outcomes} are possible. Additionally, if one considers monetary and material rewards to be relevant in the current action situation, it is just enough to introduce a {\em payoff} predicate, initialized, for example, with \mintinline[breaklines]{prolog}{initially(payoff(A,0)) :- role(A,some_role)}. Then, its evolution can be regulated with control rules, that map (possibly joint) actions to monetary gains.

\begin{table}[b]
	\caption{Syntactic restrictions for the \texttt{Condition} and \texttt{Consequence} fields for every of the proposed rule types. $\mathbf{\alpha}$ stands for an atom, i.e. a predicate symbol with terms as arguments.}
	\label{tab:syntax-by-rule-type}
	\centering
	\begin{tabular}{|c|c|c|}
		\hline \textbf{Rule type} & \textbf{\texttt{Condition}} & \textbf{\texttt{Consequence}} \\
		\hline Boundary & \texttt{agent(Ag)} & [$\sim$]\texttt{participates(Ag)} \\
		\hline Position & \texttt{participates(Ag)} & [$\sim$]\texttt{role(Ag,R)} \\
		\hline Choice & \texttt{role(Ag,R)} & [$\sim$]\texttt{can(Ag,Ac)} \\
		\hline Control & \texttt{joint\_action} & 
		\begin{tabular}{c}
			[$\texttt{consequence}_1$ withProb $p_1$,\\
			$\texttt{consequence}_2$ withProb $p_2$,\\
			...]
		\end{tabular}\\
		\hline
		\multicolumn{3}{|l|}{\texttt{joint\_action} ::= \texttt{does(Ag,Ac)} [and \texttt{joint\_action}]} \\
		\multicolumn{3}{|l|}{\texttt{consequence} ::= $\mathbf{\alpha}$ [and \texttt{consequence}]} \\
		\hline
	\end{tabular}
\end{table}

As for the information component, it is left unaddressed in this early version of ASL. As we will explain in \Cref{sec:semantics}, we will only consider a restricted version of imperfect information in the semantics of any ASL description (much like in GDL game specifications). Although introducing information constraints to limit players' observability would certainly make for an interesting extension, we do not include it here as we anticipate that it would greatly increase the complexity of the resulting formal games, potentially opening the door for incomplete information and imperfect recall games.

As advanced, the content of rule statements follows the syntax \texttt{if Condition then Consequence where Constraints}, with additional restrictions on the \texttt{Condition} and \texttt{Consequence} fields depending on the rule type. These restriction are displayed in \Cref{tab:syntax-by-rule-type}. Note that the boundary, position and choice rules all have an analogous syntax: one \texttt{agent}, \texttt{participates} or \texttt{role} predicate as the \texttt{Condition}, and \texttt{participates}, \texttt{role} or \texttt{can} as the \texttt{Consequence}, respectively. Also, their \texttt{Consequence} predicate might be preceded by the overwriting operator \verb|~|, although it only makes sense to use it with non-default rules.

In contrast, control rules may have in their condition multiple \texttt{does} predicates concatenated by the \texttt{and} operator to reflect the execution of joint actions. Their consequences, instead of a single predicate, is a list where each of its members consists of (possibly several) predicates concatenated with the \texttt{and} operator. The whole conjunction is assigned some probability with the operator \texttt{withProb}. In order for a control rule to be valid, the probability distribution over the potential consequences must be well-defined, i.e. all $p_i$ must fall in the range $[0,1]$ and must add up to unity.\footnote{If that is not the case, the game engine that interprets the rules will raise an error, and the description semantics will not be generated.}

\paragraph{Regimentation and deontic modalities}\label{par:regimentation}
In the norMAS field, the term {\em norms} often refers to constraints on the behavior of agents intended to coordinate their actions \parencite{Shoham1995}. Typically, norms are represented as a {\em deontic modality} (prohibited (P) or obliged (O), following \cite{Wright1951}). Additionally, such constraints can be {\em regimented} or not \parencite{Grossi2010}. In regimented domains (see e.g. \cite{LopezSanchez2013}), the nature of the application allows for the perfect enforcement of the desired constrains. Meanwhile, in non-regimented domains (e.g. \cite{Fagundes2016}), some monitoring and sanctioning mechanism is implemented in order to deter agents from violating the norms.

In this short section, we provide guidelines on how prohibitions and obligations on actions can be emulated through ASL rule statements, both in the regimented and sanctioning versions. We should first note that, in order for any action to be possible, it first needs to be explicitly included in a choice rule and assigned to the role capable of executing it.

Regimented constraints are addressed through choice rules, as they effectively deny the possibility to execute some action(s). For example, suppose that the default rules (that model the ``unconstrained'' situation) are to be overwritten in order to prohibit some forbidden action. Such a situation is equivalent to introducing a higher priority choice rule of the form:
\begin{flushleft}
	\mintinline[breaklines]{prolog}{rule(Id,choice,N,if role(Ag,role) then ~can(Ag,forbidden_action) where [conditions_for_prohibition]).}
\end{flushleft}
where \texttt{N} is a strictly positive integer and \texttt{role} is the role that was originally assigned the action in question.

Similarly, regimented obligation can also be modeled through choice rules. To do so, we use the following equivalence between the prohibited and obliged deontic modalities:
\begin{equation}\label{eq:deontic-equivalence}
	O\left(a_i\right) \iff \left(\forall a_j, a_j\neq a_i \implies P\left(a_j\right) \right)
\end{equation}
which states that the obligation to perform action$a_i$ is equivalent to the prohibition of performing any other action $a_j$. Then, a regimented obligation can be expressed in ASL as:
\begin{flushleft}
	\mintinline[breaklines]{prolog}{rule(Id,choice,N,if role(Ag,role) then ~can(Ag,Ac) where [Ac\=obliged_action,conditions_for_obligation]).}
\end{flushleft}

Non-regimented norms assume that it is not possible to completely ban the execution of forbidden actions, and instead they attempt to discourage agents away from them. In ASL, this approach can be expressed through control rules. A deterrence for a forbidden action follows the template:
\begin{flushleft}
	\mintinline[breaklines]{prolog}{rule(Id,control,N,if does(Ag,forbidden_action) then [punishment withProb P, no_punishment withProb 1-P] where [conditions_for_prohibition]).}
\end{flushleft}

Similarly to regimented norms, the leap from non-regimented prohibition to obligations is made thanks to \cref{eq:deontic-equivalence}:
\begin{flushleft}
	\mintinline[breaklines]{prolog}{rule(Id,control,N,id does(Ag,Ac) then [punishment withProb] where [Ac\=obliged_action,conditions_for_obligation]).}
\end{flushleft}

\subsection{Valid ASL descriptions}\label{subsec:asl-description}
Given the syntax of ASL, a valid action situation description $\mathbb{A}$ can be written as a finite set of clauses. We organize them into three exclusive subsets, one for every set of exogenous variables determining the structure of an action situation (see \Cref{fig:iadframework}):
\begin{equation}
	\mathbb{A} = \Delta \cup \Sigma \cup \Omega
\end{equation}
where:
\begin{itemize}
	\item $\Delta$ is the {\em agents base}, which includes the information on the agents and their attributes.
	\item $\Sigma$ is the {\em states base}, which includes the information on biophysical features, plus the \texttt{initially}, \texttt{terminal} and \texttt{incompatible} clauses.
	\item $\Omega$ is the {\em rules base}, which contains the \texttt{rule} statements.
\end{itemize}

In order to ensure the unambiguous interpretation of an action situation description, some limitations are placed in the use of the reserved predicates within the body of clauses. Additionally, some minor directives are intended to ensure the clear separation between the community attributes, the biophysical features and the rules into three separate knowledge bases.

\begin{definition}[valid ASL description]
	A {\em valid} ASL description $\mathbb{A} $ is a finite set of clauses split into three exclusive subsets $\Delta \cup \Sigma \cup \Omega$, where:
	\begin{itemize}
		\item \texttt{agent} appears only as the head of clauses $\Delta$.
		\item \texttt{initially}, \texttt{terminal} and \texttt{incompatible} appear only in the head of clauses in $\Sigma$. Also, \texttt{initially} clauses do not contain \texttt{can} nor \texttt{does} in their body, and \texttt{incompatible} does no have a reserved predicate symbol as its first argument.
		\item \texttt{rule} statements appear only as facts (clauses with no body) in $\Omega$, plus they follow the syntactic restrictions exposed in \Cref{tab:syntax-by-rule-type}.
	\end{itemize}
\end{definition}

\subsection{First example: Iterated Prisoner's Dilemma}\label{subsec:iterated-pd}

\begin{figure}[t]
	\centering
	\begin{subfigure}{0.4\textwidth}
		\centering
		\begin{tabular}{|r|c|c|}
			\hline & \textbf{cooperate} & \textbf{defect} \\
			\hline \textbf{cooperate} & 6,6 & 0,9 \\
			\hline \textbf{defect} & 9,0 & 3,3 \\
			\hline
		\end{tabular}
		\caption{}
		\label{subfig:pd-normal-form}
	\end{subfigure}
	\begin{subfigure}{0.35\textwidth}
		\includegraphics[width=\textwidth]{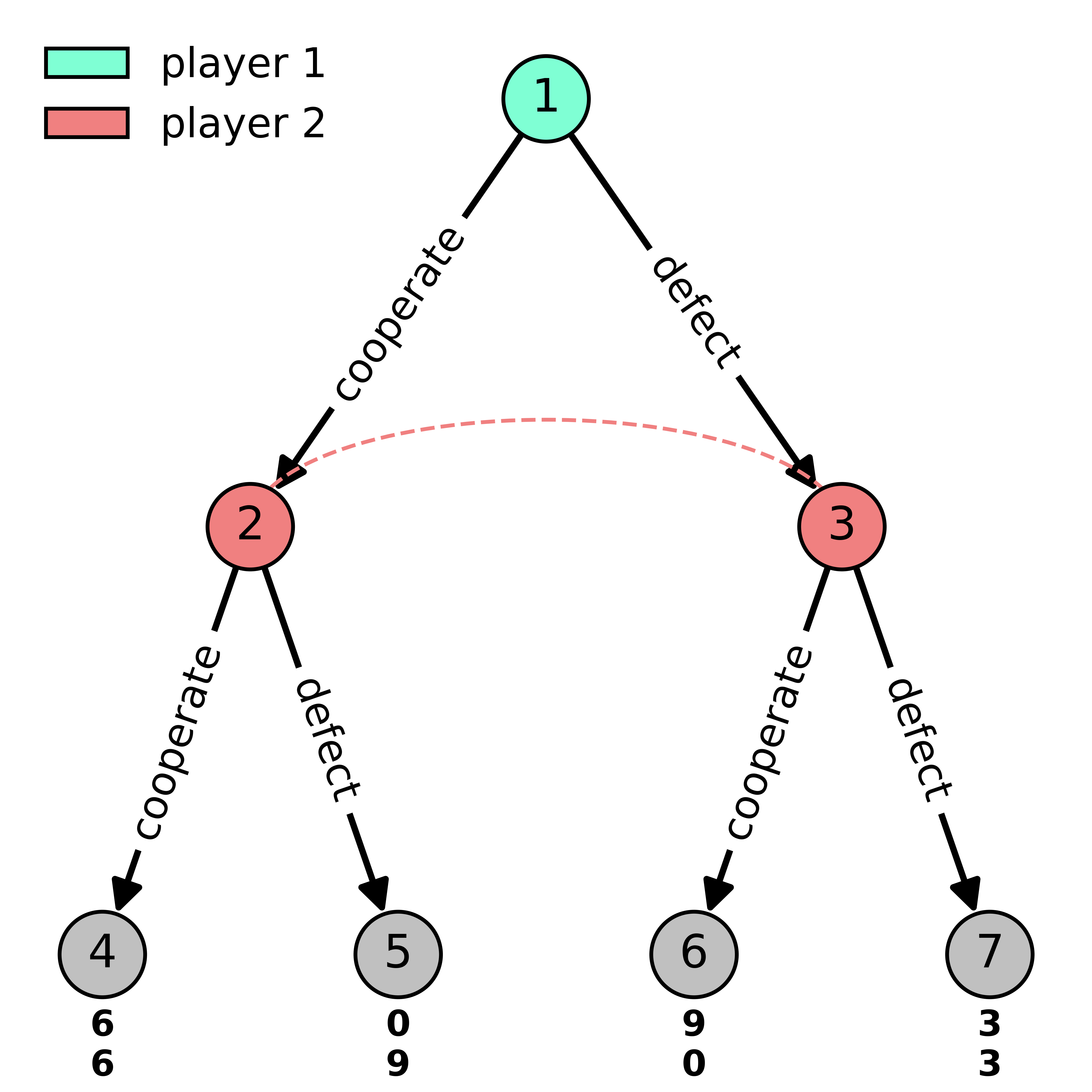}
		\caption{}
		\label{subfig:pd-extensive-form}
	\end{subfigure}
	\caption{The Prisoner's Dilemma game in the (a) normal-form and (b) extensive-form versions. Note the use of an information set for player 2 (denoted with the dashed line) to capture the simultaneous nature of the moves.}
	\label{fig:pd-normal-extensive}
\end{figure}

\begin{listing}[t]
\begin{minipage}{0.2\textwidth}
\begin{minted}{prolog}
agent(alice).
agent(bob).
\end{minted}
\end{minipage}
\hfill
\begin{minipage}{0.75\textwidth}
\begin{minted}{prolog}
initially(payoff(P,0)) :- role(P,prisoner).
initially(consecutiveDefections(P,0)) :- role(P,prisoner).
initially(rounds(0)).

terminal :- rounds(N),N>=3.

incompatible(rounds(_),L) :- member(rounds(_),L).
incompatible(payoff(P,_),L) :- member(payoff(P,_),L).
incompatible(consecutiveDefections(P,_),L) :-
	member(consecutiveDefections(P,_),L).
\end{minted}
\end{minipage}
\vspace{1.0em}
\caption{Agents $\Delta$ (left) and states $\Sigma$ (right) bases for the iterated Prisoner's Dilemma action situation, played for a total of three rounds.}
\label{code:ipd-agents-states}
\end{listing}

The best way to get a grasp on the syntax of ASL is to go through a simple illustrative example. In this section, we show how the iterated version of the benchmark Prisoner's Dilemma game can be described in ASL. Its simplicity and familiarity to a wide range of audiences makes it a perfect candidate to be the first described with our language. Other more sophisticated examples are presented in \Cref{sec:examples}.

In the Prisoner's Dilemma game, two agents are put in identical positions, where they can choose one of two actions (either ``cooperate'' or ``defect''), with no prior opportunity to communicate with one another. If an agent cheat on the other, the defector receives a large temptation payoff ($T=9$) at the expense of the sucker ($S=0$). If the two agents cooperate, they both receive identical reward payoffs ($R=6$) that are larger than the punishment reward $P=3$) they get when both choose to defect. This game has attracted a lot of interest from scholars in a wide range of disciplines because, although the most socially beneficial thing to do is for both agents to cooperate with one another, rational behavior stipulates that they should both defect, leading to worse individual and group reward. In game theoretical terms, the Nash equilibrium is not Pareto optimal.

In its iterated version, agents typically start with wealth equal to zero.They play several consecutive rounds of the one-shot game, and their wealth is increased by the reward they get at every round. The interaction is halted after some pre-defined number of rounds have been completed.

\begin{listing}
\begin{minted}{prolog}
rule(ipd,boundary,0,if agent(A) then participates(A) where []).

rule(ipd,position,0,if participates(A) then role(A,prisoner) where []).

rule(ipd,choice,0,if role(P,prisoner) then can(P,cooperate) where []).
rule(ipd,choice,0,if role(P,prisoner) then can(P,defect) where []).

% control
% P1@<P2 avoids equivalent instantiations of some control rules
rule(ipd,control,0,if does(P1,_) and does(P2,_) then [rounds(M) withProb 1] 
	where [P1@<P2,rounds(N),{M=N+1}]).

rule(ipd,control,0,if does(P1,cooperate) and does(P2,cooperate)
	then [payoff(P1,Y1) and payoff(P2,Y2) withProb 1]
	where [P1@<P2,payoff(P1,X1),payoff(P2,X2),{Y1=X1+6},{Y2=X2+6}]).

rule(ipd,control,0,if does(P1,defect) and does(P2,defect)
	then [payoff(P1,Y1) and payoff(P2,Y2) withProb 1]
	where [P1@<P2,payoff(P1,X1),payoff(P2,X2),{Y1=X1+3},{Y2=X2+3}]).

rule(ipd,control,0,if does(P1,cooperate) and does(P2,defect)
	then [payoff(P1,Y1) and payoff(P2,Y2) withProb 1]
	where [payoff(P1,X1),payoff(P2,X2),{Y1=X1+0},{Y2=X2+9}]).

rule(ipd,control,0,if does(P,defect) then [consecutiveDefections(P,M) withProb 1]
	where [consecutiveDefections(P,N),{M=N+1}]).

rule(ipd,control,0,if does(P,cooperate)
	then [consecutiveDefections(P,0) withProb 1] where []).
\end{minted}
\caption{Rule base for the iterated Prisoner's Dilemma action situation.}
\label{code:ipd-rules}
\end{listing}

First, the agents and states knowledge bases appear in \Cref{code:ipd-agents-states}. In the agents base, two agents are declared with no additional attributes. In the states base, the initial conditions are set to zero wealth for all prisoners. Their respective number of consecutive defections is also set to zero.\footnote{The \texttt{consecutive\_defections} variables will become relevant once we implement higher priority rules in this example.} Also, a counter for the number of rounds is initialized. Finally, the \texttt{incompatible} predicates state that there can only one \texttt{rounds} predicate, as well as one \texttt{payoff} and  one \texttt{consecutive\_defections} predicates per agent in a state.

Second, the rule base is displayed in \Cref{code:ipd-rules}. There are 8 rules in total: 1 boundary, 1 position, 2 choice and 6 control. They are all identified by the tag \texttt{ipd} and have priority equal to zero since they are the defaults. The boundary, position and choice rules are all very generic. All agents are allowed to participate by the boundary rules, they take on one role (that of a prisoner) according to the position rules and they are allowed to cooperate or defect with no further constraints by the choice rules.

The control rules exemplify the use of the \texttt{Constraints} field. The first one declares that, in order for the rounds counter to advance, two distinct participants must take some action. The other three control rules have identical structures as they all control the rewards received as a function of the joint actions at every stage of the game. Although there are four possibilities in the Prisoner's Dilemma game (see \Cref{subfig:pd-normal-form}), due to symmetry we need only three control rules. Finally, the last two rules regulate how the number of consecutive defections increase or are set back to zero based on whether agents cooperate or defect. There are no stochastic effects in this action situation, hence all control have one single consequence with probability equal to unity.

\section{Rule interpretation}\label{sec:rule-interpretation}
As introduced in \Cref{subsec:contributions}, an ASL description has its semantics generated as a formal game model by an automatic engine. To do so, the game engine has to repeatedly interpret the rules in place. This task is performed by the ``interpreter.pl'' script.

In this section, we go through the rule interpretation process in detail. We split it into two steps: (1) rule activation and (2) processing of consequences.  Throughout, we denote again an action situation description, split its three components, as $\mathbb{A} = \Delta \cup \Sigma \cup \Omega$. On several occasions, this set of clauses is expanded by fluents on the participants $\phi = \{\texttt{participates}(ag_1),...\}$, their roles $\rho = \{\texttt{role}(ag_1, r_1),...\}$, the current state of the system $s_t = \{f_1,...,f_n\}$ and/or the actions executed by agents $\mu = \{\texttt{does}(ag_1,ac_1),...\}$. As we will see in \Cref{sec:semantics}, the participants $\phi$ and roles $\rho$ remain constant throughout the interaction, while the joint actions $\mu$ and the current state $s_t$ change as the systems evolves.

\begin{listing}[b]
\begin{minted}{prolog}
% query_rule/1
% query_rule(?Rule): True if Rule is active given the current state of the system.
query_rule(rule(ID,Type,Priority,if Condition then Consequence where Constraints)) :-
	rule(ID,Type,Priority,if Condition then Consequence where Constraints),
	maplist(query,[Condition|Constraints]).

query(Q) :- call(Q).
query(A and B) :- query(A),query(B).
\end{minted}
\caption{Prolog interpreter predicates to find which rules are activated given the current state of the system.}
\label{code:rule-interpreter-core}
\end{listing}

First, the interpreter has to find which rules in $\Omega$ are {\em active} given the current state of the system. In general, \texttt{rule} clauses contain free variables that have to be bounded to constant given the ASL description $\mathbb{A}$ plus the suitable additional information. When a rule is fully instantiated, we say that it has been {\em activated} or {\em triggered}. A rule may be activated multiple times, as many as possible instantiations of its free variables there are given the state of the system.

The clause responsible for finding out active rules is \texttt{query\_rule(Rule)}, shown in \Cref{code:rule-interpreter-core}. Its examination reveals that, in fact, a rule statement in ASL:
\begin{flushleft}
	\mintinline{prolog}{rule(..., if Condition then Consequence where [Constraint1, Constraint2 ...]).}
\end{flushleft}
is equivalent to a traditional clause:
\begin{flushleft}
	\mintinline{prolog}{Consequence :- Condition, Constraint1, Constraint2, ... .}
\end{flushleft}

However, beyond its friendlier syntax, rule statements contain a \texttt{Type} argument, that allows to activate different kinds of rules depending on the step of the game construction process that the engine is in. Additionally, the \texttt{Priority} field helps sort out conflicts between norms with contradicting or incompatible consequences, and also filter out rules whose priority is above some threshold.

The rule activation step is common to all rule types, and it helps retrieve the consequences that rules have and that will have some effect on the resulting interaction. But first, these consequences need to be processed. Since boundary, position and choice rules all have analogous syntactic restrictions, the processing of their derived consequences is also shared., we deal with these three separately from the more sophisticated control rules.

\subsection{Processing of boundary, position and choice rules}\label{subsec:interpretation-bpc}
As displayed in \Cref{tab:syntax-by-rule-type}, boundary, position and choice rules have similar syntactic restrictions and their \texttt{Consequence} field contains a single term. The consequences for all of this rule types are processed by the function \textsc{Get-Simple-Conseqs} (see \Cref{alg:get-simple-conseqs}). First, the database is queried in order to find the active rules of the input type. The $n : f$ pairs are stored (where $n$ is the integer rule priority and $f$ its derived fluent), and those whose priority is over some input threshold are neglected.

Next, the pairs are ordered in descending order of \texttt{Priority} value (ties broken arbitrarily). Then, the fluent $f$ derived from \texttt{Conseqs} are added to the set of output ground atoms if neither that same predicate nor its overwriting (with the prefix operator $\sim f$) have already been added to the output consequences by a higher priority rule. Finally, fluents preceded by the overwriting operator are deleted before the set of facts is returned.

The rule interpreter counts on the fact that rules of the same type and priority whose consequences overwrite one another cannot be simultaneously activated. For example, the following should not be included in the database:
\begin{flushleft}
\mintinline{prolog}{rule(...,boundary,1,if agent(A) then participates(A) where [age(A,N),N>18]).}

\mintinline{prolog}{rule(...,boundary,1,if agent(A) then ~participates(A) where [age(A,N),N>18]).}
\end{flushleft}

If such rules where included, the return fluent set of \textsc{Get-Simple-Conseqs}(..., boundary, 1) would not be deterministic, as it would depend on how the ties between the two conflicting rules are broken.

Note that the sole ASL description $\mathbb{A}$ is enough only in the case of processing the boundary rules. The set of participants $\phi = \{\texttt{participates}(ag_1),...\}$ should be added to the action situation description before processing the consequences of position rules. Likewise, in order to process the consequences of choice rules both the roles $\rho = \{\texttt{role}(ag_1, r_1),...\}$ and the current state $s_t = \{f_1,...,f_n\}$ have to be added to $\mathbb{A}$. These datum are necessary to ensure the proper activation of the various rule types.

\subsection{Processing of control rules}\label{subsec:interpretation-control}
Control rules have quite distinct syntax with respect to the other types, hence their consequences are processed by a different function, \textsc{Get-Control-Conseqs} (see \Cref{alg:get-control-conseqs}). The added difficulty arises from the fact that possibly joint actions have, in general, stochastic consequences. In turn, every potential consequence does not just correspond to a single predicate, but to several. Thus, it is not enough to simply return a set of fluents, but rather a set of potential next states and a probability distribution over those, where every state is in turn a set of fluents that characterize it.

As control rules regulate how the state of the system transitions due to the actions performed by the agents, the pre-transition state $s_t = \{f_1,...,f_n\}$ as well as the action fluents $\mu = \{\texttt{does}(ag_i,ac_i),...\}$ need to be added to the ASL description $\mathbb{A}$. Additionally, it must hold that:
\begin{equation}\label{eq:single-action-assumption}
	\forall ag_i \mid \exists \texttt{does}(ag_i,ac_j) \in \mu \implies |\{ac_j \mid \texttt{does}(ag_i,ac_j) \in \mu\}| = 1
\end{equation}
meaning that any agent may take at most one action at a time.

The processing of control rule consequences starts off just as that of the other rule types. First, the instantiations of activated control rules are stored, together with their priority as a key (excluding those exceeding some input threshold) and sorted by descending priority. The set of potential next states $\mathcal{S}_{t+1}$ is initialized as an set of sets and unity priority assigned to the empty set.

Then, the function loops over the activated rules. Every rule is composed of a probability distribution over joint consequences, which in turn contain several fluents concatenated by the \texttt{and} operator. The following check is performed. If any of the single fluents in any of the potential joint consequences is found to be incompatible by any of the provisional next states in $\mathcal{S}_{t+1}$, then that rule, despite having been activated, is ignored. This is done in order to avoid inconsistencies in the final probability distribution over the post-transition states.

If an activated control rule passes the check, its consequences are added to the set of potential next states. Every provisional post-transition state $s_{t+1} \in \mathcal{S}_{t+1}$ is expanded with the fluents from each of the joint consequences of the control rule being examined. The probability of the new expanded state is updated as the product between the probability of $s_{t+1}$ prior to expansion times that of the joint consequences, provided by the active control rule.

One final step is performed before returning the post-transition states and their probability distribution. A loop is run over the pairs of pre-transition state fluents and the post-transition states. If the pre-transition state fluent is found to be compatible with the facts in the provisional post-transition state, it is dragged over and added to the potential next state. When this loop is complete, the set of post-transition state $\mathcal{S}_{t+1}$ alongside with the probability distribution over those is returned. Note that, because of this final step, if we were to call \textsc{Get-Control-Conseqs} but no control rules were activated, the function would just return the pre-transition state ($\mathcal{S}_{t+1} = \{s_t\}$) with unit probability ($\mathcal{P}(s_t) = 1$).

The use of the \texttt{incompatible} predicate together with the dragging of pre-transition state fluents is the approach that ASL takes to one of the problems that any action formalism has to inevitably address: the {\em frame problem} \parencite{Lin2008}. The frame problem states that when the state of an action is axiomatized, it should not be necessary to refer to the facts that are not affected by it. In ASL, control rules need only include the facts that do change in their \texttt{Consequence} field. By introducing the \texttt{incompatible} predicate and having the interpreter drag the old fluents that are consistent with those newly derived, the ASL allows for natural expression of control rules where only the facts affected by the actions included in the \texttt{Condition} field need to be stated.

\section{Language semantics}\label{sec:semantics}
As reiterated throughout the test, an action situation description has its semantics grounded as an EFG with a restricted use of imperfect information. Furthermore, the formal game representation is augmented with a set of fluent at those tree nodes that directly correspond to {\em states} of the system. In order not get ahead of ourselves, we first need to review some common definition from Game Theory and complement them with some definitions unique to this work. We then take deep dive at how a formal game is built from arbitrary action situation description, and which properties it is ensured to have as a consequence of the building mechanism.

\subsection{Background on Game Theory}\label{subsec:game-theory}
In the game theory field, the simplest model of a multiagent interaction is provided by normal-form games:
\begin{definition}[normal-form game]
	A {\em normal-form game} is a tuple:
	\begin{equation*}
		G=\left(P,(A_i)_{i \in P},(U_i)_{i \in P}\right)
	\end{equation*}
	where:
	\begin{itemize}
		\item $P$ is the set of {\em players} (or agents).
		\item $A_i$ is the set of {\em actions} available to player $i$.
		\item $U_i: A \rightarrow \mathbb{R}$ is the {\em utility function} for player $i$, which maps every possible joint action profile in $A = \bigtimes\limits_{i \in P} A_i$ to a numeric quantity.
	\end{itemize}
\end{definition}

Normal-form games have been widely studied in a variety of fields, from microeconomics to evolutionary theory. However, they are not suitable to capture sequential interactions where agents might take several actions at different times. For this reason, normal-form games are sometimes referred to as stateless games. Moreover, normal-form games do not explicitly store information on possible stochastic effects of joint actions.

To address these shortcoming, it is necessary to work with extensive-form games, where the strategic interaction is represented as tree that agents transverse as they take actions. To account for stochastic effects, a new artificial player $p_0$ is added to the players set, which is usually referred to as ``nature'' or ``chance''.

\begin{definition}[extensive-form game]
	An {\em extensive-form game} is a tuple:
	\begin{equation*}
		G=\left(P,(X,E),T,W,\mathscr{A},p,U\right)
	\end{equation*}
	where:
	\begin{itemize}
		\item $P$ is the set of {\em players} (or agents).
		
		\item $(X,E)$ is the {\em game tree}, where $X$ is the set of nodes (or vertices ) and $E \subseteq X \times X$ is the set of directed edges. One of the nodes $x_0 \in X$ is the {\em root} of the tree (with no incoming edges), such that for all other nodes $x \in X\setminus \{x_0\}$ there is a unique path from $x_0$ to $x$. The subset of nodes $Z \subseteq X$ with no outgoing edges are called the {\em terminal} (or {\em lead}) nodes.
		\item $T: X \setminus Z \rightarrow P \cup \{p_0\}$ is the {\em turn function}, which assigns every non-terminal node to the player responsible for taking an action at that node, including chance moves. The turn function induces a partition $\Theta = \{\Theta_0,\Theta_1,...,\Theta_{|P|}\}$ over the non-terminal nodes, by sorting them into subsets according to the player whose turn it is:
		\begin{equation*}
			\Theta_i = \{x \in X \setminus Z \mid T(x) = i\}
		\end{equation*}
	
		\item $W=\{W_i\}_{i \in P}$ is the {\em information partition} which, for every player $i \in P$, $W_i$ corresponds to a partition of $\Theta_i$. $W_i$ splits all the nodes where $i$ makes a move into mutually exclusive sets. Every $w_i \in W_i$ is called an {\em information set}. When player $i$ makes a move, the information available to $i$ is exactly the same at any of the nodes that belong to the same information set.

		A game where all information sets contain just one node are called {\em perfect information games}. In this case, a player always knows precisely what node it is making the move from. When that is not the case, games are said to have {\em imperfect information}. Although information partitions are typically interpreted as partial observability, they can also be employed to simulate simultaneous moves. The prototype for this case can be found in the extensive form of the one-shot Prisoner's Dilemma game (see \Cref{fig:pd-normal-extensive}).
		
		\item $\mathscr{A} = \left(A(w)\right)_{w \in W}$ denotes the {\em actions} available to the players, where $A(w)$ is the set of actions available at information set $w$ for the player whose turn it is to move at that set. Actions serve as labels on the outgoing edges from a decision (non-chance) node.
		
		\item $p$ is a function that assigns to every chance node $x \in \Theta_0$ a probability distribution over its outgoing edges. Hence, it describes the nature of the environment and its stochastic effects.
		
		\item $U= \left(U_i\right)_{i \in P}$ is the {\em utility} function which assigns, for every agent, a numerical payoff for every terminal node, $U_i: Z \rightarrow \mathbb{R}$.
	\end{itemize}
\end{definition}

EFGs provide much richer representations of multiagent interactions. To generate the semantics of an ASL description, we will use a restricted form of EFGs to model all the possible ways by which a state $s_t$ can evolve given the actions available to the participants at that state. We name this restricted form of EFGs as {\em game rounds}:

\begin{definition}[game round]\label{def:game-round}
	A {\em game round} is an extensive-form game with the following characteristics:
	\begin{enumerate}
		\item The root node is never a chance node.
		\item There is, at most, one information set per player.
		\item Any two nodes $x_1$, $x_2$ that belong to the same information set, the length from the root to $x_1$ and from the root to $x_2$ must be equal.
		\item If $T(x)=p_0$ ($x$ is a chance node), then all of its child nodes are terminal.
	\end{enumerate}
\end{definition}

In practice, in a game round every player has the opportunity to make only one move. Every depth level corresponds to the information of one player (excluding the terminal nodes and possibly their immediate parents). In a game round, every path of play corresponds to a joint action, and the sequence of player whose turn it is to take is constant across paths. If the joint action has deterministic results, then the decision vertices are succeeded by a leaf node. Otherwise, if the joint action has stochastic effects, then a chance node succeeds the last decision node, before leading to a leaf node. Note than condition (3) ensures that game rounds are always {\em perfect recall}. This requirement will greatly facilitate the equilibria computations later on.

For example, the game in \Cref{subfig:pd-extensive-form} holds the requirements to be a game round. Pointing this example raises the following question: why have we not chosen to model the interaction at a single time-step as a normal-form game, instead of going to the extent of using the much more loaded extensive form and then restrict it?

The answer is totally motivated by the ability of EFGs to explicitly store the information on probabilistic effects. This is not possible in normal-form games, where every action profile is mapped to a single payoff vector. Whether this payoff is a weighted average over multiple possible outcomes is not discernible from a normal-form representation. In our opinion, the loss of this information rules out normal-form games as suitable representations for action situation interactions.

Given a state of the world, a game round models all the possible ways by which, through a single action each, agents might alter the state. Since we want to model relatively complex action situations, with agents executing actions not once but multiple times, several game rounds will be concatenated and merged into a larger extensive-form game. Next, we explain how game rounds are built from an action situation description and merged into an extensive-form game that grounds the semantics of the description.

\subsection{From action situation descriptions to games}\label{subsec:build-games}
Now, we move on to the process that takes an ASL description as input and returns its game semantics as output. This process is performed by the iterative generation of single game rounds plus their concatenation into a larger game structure.

First, we address how a single game round is built by the function \textsc{Build-Game-Round} (see \Cref{alg:build-game-round}). It starts off with the set of facts that characterize the pre-transition state $s_t$, which acts as the root of the round. Then, the game tree is built in a {\em breadth-first} manner, by iterating over the agents that are able to take some action at $s_t$ according to the choice rules. One information set is added for each of them. At every iteration of this loop, the depth of the game round tree is increased by one level.

Once the tree skeleton has been built, it is time to find out which fluents hold true at the potential post-transition states $s_{t+1} \in \mathcal{S}_{t+1}$. Every terminal node corresponds to a distinct joint action executed from $s_t$. Then, for every terminal node the joint action that leads to it from the root node is retrieved and added to the action situation description $\mathbb{A}$. Next, \textsc{Get-Control-Conseqs} is called in order to find out the post-transition states that may be reached from that action profile. In case there are no stochastic effects ($|\mathcal{S}_{t+1}| = 1$) the fluents of the sole next state are assigned to the terminal node. Otherswise, the terminal node is converted into a chance node, and additional descendants are added, one for every potential next state $s_{t+1} \in \mathcal{S}_{t+1}$. The fluents at these new terminal nodes and the probabilities of the edge from the parent chance node are set as the return values of \textsc{Get-Control-Conseqs}. Finally, the joint actions are deleted from the database before moving on to the next terminal node.

By construction, the extensive-form game returned by \textsc{Build-Game-Round} fulfills the properties of a game round according to \Cref{def:game-round}. Note that in every game round built by this function, only the root and terminal nodes correspond to actual states of the system, and they are assigned the fluents that hold true at that state (in fact the fluents that hold true at the root node are assigned before the tree emanating form it is constructed). The intermediate nodes between the root and the terminal ones do no correspond to actual states of the system. Hence, they are not assigned any fluents. In fact, they are auxiliary nodes, whose function is to capture the simultaneous nature of joint moves and the possibility for random effects in the environment.

\textsc{Build-Game-Round} manages one last important point. For every joint action profile, it checks, for every joint action $\mu$ that might be available at $s_t$, whether its execution leads to termination. If that is the case, the terminal node that corresponds to the execution of $\mu$ in state $s_t$ will not be considered for further expansion (i.e. the construction of the game rounds that emanates from it) when the complete game for the action situation is built.

Now that we know how to model the possible ways by which a state of the system $s_t$ might evolve, the only step that is left is to concatenate multiple rounds into a larger game tree. This is precisely what the function \textsc{Build-Full-Game} does (see \Cref{alg:build-full-game}). It only needs an action situation description as data, and maintains a queue of states susceptible to be expanded. The queue is initialized with the initial state of the interaction, which is derived as the instantiations of the query \texttt{initially(Fact)}, {\em after} the participants and their roles have been added to the database by processing the boundary and control rules respectively. As new game rounds are built, their respective terminal nodes are pushed to the queue as long as the joint actions leading to them from the root node in their game round do not trigger termination.

Additionally, every time a state is popped from the queue and its fluents added to the ASL description, \textsc{Build-Full-Game} checks whether the termination conditions hold, {\em prior to the execution of any action}. Adding this check to the one performed in \textsc{Build-Game-Round} shows that, in an ASL description, a state may lead to termination either because the facts that characterize it command it, or because of the actions that have led to it from its pre-transition state.

Finally, \textsc{Build-Full-Game} returns the extensive-form game that represents the action situation description, augmented by the facts that characterize the nodes that directly correspond to states of the system (those that are the root of game rounds plus the leafs). Hence, we can define the semantics of a valid ASL description by construction:

\begin{definition}[ASL semantics]
	The {\em semantics of a valid ASL description} $\mathbb{A}$ correspond to an extensive-form game with a restricted use of imperfect information $\Gamma$, and a function $\mathcal{F}$ over a subset of the nodes in $\Gamma$, both of which are the return values of \textsc{Build-Full-Game}. $\Gamma$ is built as the concatenation of game rounds, and is augmented with $\mathcal{F}$, which returns a set of fluents for every node in $\Gamma$ that correspond to a proper states of the system.
\end{definition}

It should be noted that \textsc{Build-Full-Game} does not deal with the {\em utilities assigned to the leaf nodes} of the game tree. We choose not to assign utilities at construction time in order to allow flexibility in the valuations that agents make of outcomes. If one is conducting a classical game-theoretical analysis based on material rewards, a predicate standing for such variable can be introduced and its evolution modeled through control rules, as stated in \Cref{subsec:rules-syntax}. Then, for every agent, the utility at a terminal node is extracted from the \texttt{payoff(agent,$x$)} fluent assigned to that node. Alternatively, if one wishes to model other values that they agents might take into consideration in the particular action situation, the utilities can be assigned as a function of the fluents that hold true at every terminal node and/or the path of play from the root to the terminal node.

\subsection{Follow-up on the iterated Prisoner's Dilemma}\label{subsec:example-ipd-semantics}
As an example of ASL semantics generation, we go back to the iterated Prisoner's Dilemma example first presented in \Cref{subsec:example-ipd-semantics}. The extensive form game built from the default rules presented there is shown in \Cref{fig:ipd-default} (see \Cref{sec:ipd-game-trees}). The utilities that appear below the terminal nodes have been set by the \texttt{payoff(Agent,X)} fluents assigned to it.

To illustrate the introduction of higher priority rules and the impact these have on the semantics, we propose two examples. First, we consider a ban on the number of consecutive defections that agents can take. Second, we introduce random changes to the outcome resulting from both agents playing ``defect''.

\begin{listing}[b]
\begin{minted}{prolog}
% rules to limit the number of consecutive defections
rule(ipd,choice,1,if role(P,prisoner) then ~can(P,defect)
	where [consecutiveDefections(P,N),N>=2]).

% rules to change the outcome of mutual defection
rule(ipd,choice,2,if role(P,prisoner) then can(P,defect) where []).

rule(ipd,control,2,if does(P1,defect) and does(P2,defect)
	then [payoff(P1,Y11) and payoff(P2,Y12) withProb 0.5,
	payoff(P1,Y21) and payoff(P2,Y22) withProb 0.5]
	where [P1@<P2,payoff(P1,X1),payoff(P2,X2),{Y11=X1+0},{Y12=X2+9},{Y21=X1+9},{Y22=X2+0}]).
\end{minted}
\caption{Additional regulations with priority 1 to limit the number of consecutive defections (top), and with priority 2 to change the outcome of the mutual defection.}
\label{code:ipd-additional-rules}
\end{listing}

The additional rules needed to introduce those regulations appear in \Cref{code:ipd-additional-rules}. At the top, in order to limit the number of consecutive defections, it is enough to introduce one new choice rule with priority 1 that turns the action ``defect'' unavailable if the agent in question has defected twice in a row. This is an example of a prohibition rule in the regimented modality. The EFG that results from the interaction of this rule appears in \Cref{fig:ipd-limit-defections}. Now, at some game rounds some agents only have action ``cooperate'' at their disposal, and the symmetry of the game tree is broken.

At the bottom, the rules that change the outcome of the mutual defection are introduced with priority 2. First, they need to undo the ban on several consecutive defection through a control rule that recovers the action ``defect'' under any circumstances. Then, we present the control rule that is most representative of this regulation. Now, if both agents defect, the resulting outcome is {\em as if} one agent had cheated on the other. The selection for who becomes the {\em de facto} cheater and sucker is done by flipping an unbiased coin. The resulting game game semantics for this configuration are displayed in \Cref{fig:ipd-ban-mutual-defect}.

\section{Implementation and computation of equilibria}\label{sec:implementation}
Now that we have extensively presented the formal syntax and semantics of ASL, we go into the practical aspects of writing an action situation description, generating the extensive-form game that derives from it and evaluating its structure.

\begin{figure}[t]
	\centering
	\includegraphics[width=1.0\linewidth]{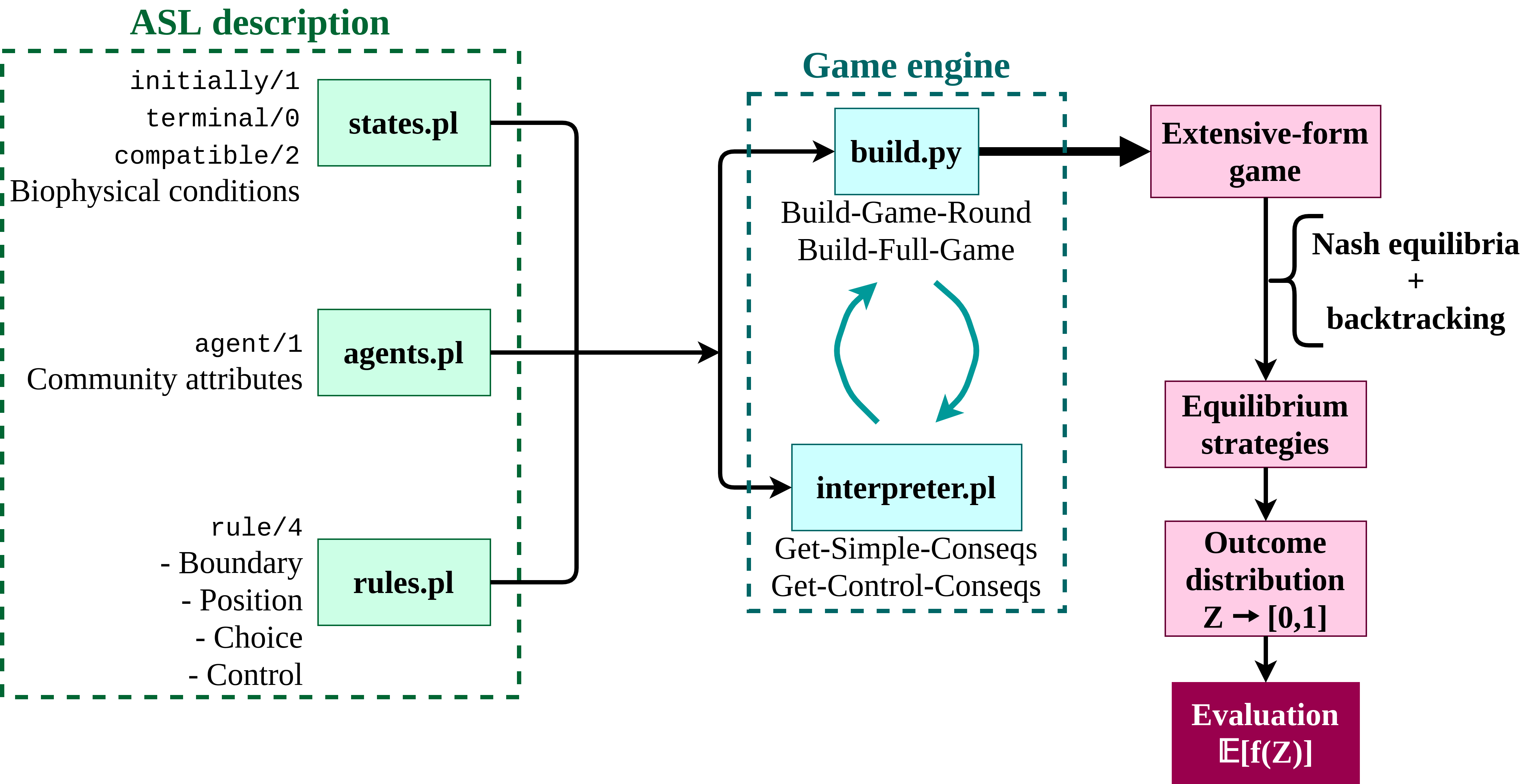}
	\caption{Implementation of the computational model of the IAD framework.}
	\label{fig:implementation}
\end{figure}

The complete set-up to perform the {\em what-if} of a rule configuration is displayed in \Cref{fig:implementation}. First, we the ASL description with its clauses divided according to the separation $\mathbb{A} = \Lambda \cup \Sigma \cup \Omega$ into agents, state-related information, and rules (see \Cref{subsec:asl-description}) into three corresponding files states.pl, agents.pl and rules.pl.

These three files are fed to the game engine, that consists of the rule interpreter and the game builder. The first is a Prolog script directly responsible for querying and processing the activated rules, and contains the implementations of \textsc{Get-Simple-Conseqs} and \textsc{Get-Control-Conseqs}. The second is a Python script that repeatedly communicates with the rule interpreter\footnote{The communication between the Prolog script interpreter.pl and the Python script build.py is realized thanks to the open-source PySwip package (\url{https://github.com/yuce/pyswip}).} and generates the action situation semantics through its implementation of the \textsc{Build-Game-Round} and \textsc{Build-Full-Game} functions.

Once the EFG is generated and utilities assigned to its terminal nodes, an assessment of its incentive structure can be performed with standard game-theoretical tools. In particular, the construction of EFGs as a concatenation of game rounds greatly facilitates their integration with {\em sequential rationality} solution concepts. Recall from \Cref{subsec:game-theory} that every game round is ensured to have perfect recall. Consequently, the resulting game tree also has perfect recall. A classic result in game theory \parencite{Kuhn1953} establishes that in imperfect-information EFGs of perfect recall, for every {\em behavioral} strategy\footnote{In an EFG, a behavioral strategy of player $i$ fixes, for every information set $w \in W_i$, a probability distribution over the available actions at $w$.} there is a corresponding equivalent {\em mixed} strategy,\footnote{In an EFG, a mixed strategy for player $i$ corresponds to a probability distribution over the sequences of actions that player $i$ as he transverses the game tree.} and vice-versa \parencite{Kuhn1953}.

Considering this result, we propose to use {\em backtracking} to compute the equilibrium strategies in an EFG generated from an ASL description, by following the steps:
\begin{enumerate}
	\item Take advantage of the fact that the game rounds that are closest to the leaf nodes are also the smallest {\em subgames} within the tree. Turn these game rounds into {\em normal-form} games, by substituting the chance nodes (if any) with terminal with utility averaged over its children.
	\item Compute the equilibrium strategies at the resulting normal-form game, using whichever solution concepts one deems appropriate. The most common notion of rationality in game theory is expressed by the Nash equilibrium, however other options are possible (e.g. correlated equilibria).
	\item Substitute the whole game round by a terminal node. Set its utilities to the average of the probability distribution induced by the previous equilibrium strategies together with the probabilities over chance nodes.
	\item Repeat until the root of the original EFG is reached.
\end{enumerate}

Note that the resulting equilibrium strategies computed by this procedure are given as {\em behavioral strategies}. For every game round, player $i$ is assigned some probability distribution over the actions available to him at that information set. Equivalently, the same probability distribution can be attributed not to the game round itself, but to the sole information set that player $i$ has within that game rounds. Consequently, the backtracking procedure assigns, for every information set in the game tree, one probability distribution for the actions available at that information set. This is precisely the definition of a behavioral strategy. However, following \cite{Kuhn1953}, we are guaranteed there is some equivalent mixed strategy.

Note also that the resulting equilibria strategies are not just rational but also {\em sequentially rational}. They correspond not just to equilibria over the whole game tree, but their restriction to every subgame is also an equilibria over it. this property is referred to as {\em subgame perfect equilibria}.

So actually, limiting the use of imperfect information to simultaneous moves (hence ensuring perfect recall) does limit the expressive power of our language, but has major advantages when it comes to computing the equilibria that the game structure incentivizes. First, only ``small'' game have to be converted from the extensive to the normal form. although this transformation can lead to exponential blow-up in the general case, we keep this complexity under control since it is game rounds (where players act just once) that have to be converted.

Second, instead of needing one big transformation from the whole EFG to a normal-form representation to find the mixed equilibria strategies directly, we can compute the equivalent behavioral strategies by performing more less intensive transformations over the game rounds only. An added advantage is that this approach avoids the redundancies often generated when turning a complex EFG into its normal-form representation, and also returns those equilibria that are sequentially rational.

The resulting equilibria strategies are available, it is straightforward to find the probability distribution they induce over the terminal nodes $Z$, together with the probabilities assigned to chance moves (see ``Outcome distribution'' box in \Cref{fig:implementation}). Finally, an evaluation criteria (which we generically denote as a function $f$ over the leaf nodes) is defined, and its expected value calculated given the induced outcome distribution. This simple computation concludes the process from a rule configuration (an ASL description) to its evaluation. Now, the communities of agents involved can quantify whether the rules in place are in accordance with their idea of a ``good'' outcome.

\section{Further examples}\label{sec:examples}
Before presenting the final conclusions, we illustrate the versatility of the ASL language with two more examples. These are more sophisticated than the very generic Prisoner's Dilemma, and also more interesting from the policy analysis and economics perspective. 

\subsection{Axelrod's (meta)-norms game}\label{subsec:axelrod-metanorms-example}
First, we use ASL to model the norms and metanorms games, originally proposed by \cite{Axelrod1986}. This action situation can be seen as a more elaborated version of the Prisoner's Dilemma. In the {\em norms} game, an individual $i$ has the opportunity to defect and get benefit at the expense of others. If he does not defect, no benefits or costs are incurred by anyone. If he does defect, then there is a fixed probability that he will be detected by a monitor $j$. If the cheater is defected, then the monitor can choose to punish him at a cost to himself, but also imposing a large loss to the cheater.

In an extension called the {\em metanorms} game, the monitor $j$ is himself being watched by a meta-monitor $k$. Similarly to the norms game, the meta-monitor may detect with some probability if the monitor has neglected his duty (i.e. has chosen not to sanction $i$ despite detecting his defection). In that case, the meta-monitor may decide whether to punish the monitor or not.

The norms and metanorms games are interesting examples for two reasons. First, they illustrate non-regimentation, as ``bad'' actions (defection by an individual or neglect by the monitor) always remain feasible actions. Furthermore, the agency of those responsible for monitoring and sanctioning is explicitly introduced into the game, instead of being idealized away as part of the environment dynamics (as in e.g. \cite{Fagundes2016}).

The complete ASL description for this action situation appears in \Cref{code:metanorms-description} (see \Cref{sec:example-ASL-descriptions}). The rules structuring the {\em norms} game are designated as the default ones and hence have priority zero. The meta-monitoring is introduced with additional regulations of priority 1. Their resulting game and state fluent semantics appear in \Cref{fig:axelrod-semantics}. The norms and metanorms game trees are constructed with \textsc{Build-Full-Game} using arguments \textit{threshold}=0 and \textit{threshold}=1 respectively. The utilities that appear below the terminal nodes are set for the agent in the first argument of predicate \texttt{payoff} to the value in the second argument.

The state fluents for the nodes that correspond to actual states of the systems are also shown. For the terminal nodes, the induced probability distribution induced by the equilibrium strategies given the utilities appear next to their fluents. The equilibrium strategies have been computed by applying the optimization approach of \cite[p. 104]{Shoham2014} in order to compute the Nash equilibria at every game round, and then backtracking the utilities.

\subsection{Ostrom's fishing game}\label{subsec:fishing-game}
The last example for which we write an ASL description first appeared in \cite[Ch. 4]{Ostrom1994}. It illustrates a theoretical analysis of how community-crafted rules are capable of transforming the opportunity structure that individuals find themselves in within common-pool resource environments. In the default situation, two fishers have access to an open-water fishery. Starting at the shore, they may go to one of two fishing spots (one is assumed to be more productive than the other). If, after the first trip, both fishers meet at the same spot, they can choose to stay or leave. If they meet yet again at the same spot after that second action, they inevitably fight. The winner of the fight is determined by flipping a biased coin, whose probabilities depend on the strength of every fisher.

The ASL description for the default situation is displayed in \Cref{code:fishing-game-description-default}. This is the first example that contains agent attributes (\texttt{speed} and \texttt{strength}) as well as biophysical conditions (two fishing spots declared with predicate \texttt{fishing\_spot}). The boundary and position rules are analogous to the other examples. Initially, both agents start at the shore. The interaction halts when each fisher is in a different spot, or when one of them has lost a fight. At every state, one fisher may only be at one place, and there can only be one loser of fights or races (more on that when we turn to the next rule configuration).

The choice rules are also very easy to interpret. They state that fishers at the shore can go to any of the fishing spots. Once there, they may choose to stay or leave. As for the control rules, they assume that fishers always get to wherever they intended to go (boats never break down). The last control rule states that, if fishers are at the same fishing spot and they both take the same action, a fight will ensue (since they meet again either at the same spot or at another one). The semantics of this ASL description (using the rules with priority equal to zero) are displayed in \Cref{fig:fishers-default-semantics}.

In order to avoid violent fights, additional higher priority rules can be introduced. The first option is to implement a {\em first-in-time, first-in-right} rule configuration. Here, fishers race from the shore to the fishing spot they go to. The first to get there is guaranteed to keep the spot, while the slower fisher has to leave. The winner of the racing contest is not determined by their \texttt{strength}, but by their \texttt{speed} instead. The additional rule statements necessary to implement this rule configuration appear in \Cref{code:fishing-game-description-race}, and the corresponding semantics in \Cref{fig:fishers-race-semantics}.

Another possibility is the implementation of a {\em first-to-announce, first-in-right} rule configuration. In this case, one of the participating agents is randomly assigned to a new role, which we refer to as {\em announcer}. Before anyone has left the shore, the announcer has to broadcast the spot where he intends to fish. Then, if he holds his promise and goes there, he is guaranteed it, i.e. he always wins the race to get there. If the announcer goes to a different spot than the one he has broadcast, a race ensues between the two fishers, analogous to the {\em first-in-time, first-in-right} configuration.

The necessary rule statements to implement this configuration appear in \Cref{code:fishing-game-description-announce}. Note that rules have priority 2, since they are added {\em on top of} the rules for the previous {\em first-in-time, first-in-right} configuration. This is the first example that illustrates a non-default position rule, such as the one that creates the {\em announcer} role. The semantics for this rule configuration appear in \Cref{fig:fishers-announce-semantics}.

The utilities set at the terminal nodes of \Cref{fig:fishers-default-semantics}, \ref{fig:fishers-race-semantics} and \ref{fig:fishers-announce-semantics} are set by assigning the following costs and benefits to actions and outcomes: a fisher keeps a spot to himself or wins the fight over it ($v_1=10, v_2=5$), a fisher looses a fight ($d=-6$), a fisher travels between spots ($c=-2$). The computation of the equilibrium strategies is identical to the one performed in \Cref{subsec:axelrod-metanorms-example}.

We evaluate the resulting outcome distributions qualitatively. For the default rule configuration, violent outcomes are predicted for over 50\% of the paths of play (terminal nodes 14 through 27). Both the {\em first-in-time, first-in-right} and the {\em first-to-announce, first-in-right} configurations avoid violence, and hence promote more socially desirable outcomes. However, for the {\em first-in-time, first-in-right} configuration the fishers still go to the same spot and compete for it. This outcome is not promoted in the {\em first-to-announce, first-in-right} configuration. Now, the announcer prefers to announce the most productive spot (spot 1) and remain faithful to his announcement. The other fisher, then, chooses to for the other spot. Hence, honest announcements are encouraged, while all types of competition are prevented with the {\em first-to-announce, first-in-right} rule configuration.

\section{Conclusions}\label{sec:conclusions}
In this work, we have presented a complete computational model of Elinor Ostrom's Institutional Analysis and Development framework. It includes the Action Situation Language, a novel logical language for the description of action situations, whose friendly syntax is highly tailored to the components identified in the IAD theory. ASL is complented with a game engine that generates the semantics of a description as an extensive-form game. The resulting EFG has some self-imposed limitations, however these facilitate greatly the computational of equilibrium strategies and the prediction of outcomes. We have illustrated ASL and the complete model with some examples. In particular, the fishers action situation in \Cref{subsec:fishing-game} demonstrates the complete pipeline, from a rule configuration to an evaluation of the outcomes incentivized by it. Additionally, it shows how additional suitable regulations are able to steer the system towards more desirable end states.

The work presented here has several limitations snd could be extended in several directions, namely:
\begin{itemize}
	\item The incorporation of {\em information rule types} that could greatly enhance the expressive capabilities of ASL, as it would potentially provide the syntax for games with imperfect information (beyond simultaneous moves), or games with imperfect recall.
	\item The possibility for dynamic boundary and position rules that are queried not just before the interaction kicks off, but also while it is ongoing. Such a refinement would let agents get in and out of an action situation and/or switch roles dynamically.
	\item Studies on the relationship between ASL and logical formalisms such as Situation Calculus. For example, how should ASL rule statements be translated into Situation Calculus domain axiomatizations?
	\item Definition of the conditions that a set of rule statements should fulfill in order to actually have an impact on the resulting semantics. For example, a position rule assigning a participant to an agent with no actions available has no effect on the outcomes of the interaction. Interesting work could be developed to define this sort of mutually dependent rule statements.
	\item Studies on the nesting of action situations. ASL descriptions can be written to express a {\em collective-choice} action situation (e.g. a negotiation domain or a voting procedure), where outcomes are linked to the adoption of new rule statements, which in turn configure the structure of an {\em operational} action situation (like the examples presented in this paper). The linkages between two such ASL descriptions are also worth exploring.
\end{itemize}

\printbibliography[heading=bibintoc]

@Article{Axelrod1986,
  author    = {Robert Axelrod},
  journal   = {American Political Science Review},
  title     = {An Evolutionary Approach to Norms},
  year      = {1986},
  number    = {04},
  pages     = {1095--1111},
  volume    = {80},
  doi       = {10.2307/1960858},
  publisher = {Cambridge University Press ({CUP})},
}

@Article{Cole2017,
  author    = {Daniel H. Cole},
  journal   = {Journal of Institutional Economics},
  title     = {Laws, Norms, and the Institutional Analysis and Development Framework},
  year      = {2017},
  number    = {4},
  pages     = {829--847},
  volume    = {13},
  doi       = {10.1017/s1744137417000030},
  publisher = {Cambridge University Press ({CUP})},
}

@Article{Fagundes2016,
  author    = {Moser Silva Fagundes and Sascha Ossowski and Jes{\'u}s Cerquides and Pablo Noriega},
  journal   = {International Journal of Approximate Reasoning},
  title     = {Design and evaluation of norm-aware agents based on Normative Markov Decision Processes},
  year      = {2016},
  pages     = {33--61},
  volume    = {78},
  doi       = {10.1016/j.ijar.2016.06.005},
  publisher = {Elsevier {BV}},
}

@InCollection{Frantz2013,
  author    = {Christopher Frantz and Martin K. Purvis and Mariusz Nowostawski and Bastin Tony Roy Savarimuthu},
  booktitle = {Lecture Notes in Computer Science},
  publisher = {Springer Berlin Heidelberg},
  title     = {{nADICO}: A Nested Grammar of Institutions},
  year      = {2013},
  pages     = {429--436},
  doi       = {10.1007/978-3-642-44927-7_31},
}

@Article{Ghorbani2016,
  author    = {Amineh Ghorbani and Giangiacomo Bravo},
  journal   = {International Journal of the Commons},
  title     = {Managing the commons: a simple model of the emergence of institutions through collective action},
  year      = {2016},
  number    = {1},
  pages     = {200--219},
  volume    = {10},
  doi       = {10.18352/ijc.606},
  publisher = {Ubiquity Press, Ltd.},
}

@InBook{Lin2008,
  author    = {Fangzhen Lin},
  chapter   = {16},
  editor    = {Frank van Harmelen and Vladimir Lifschitz and Bruce Porter},
  pages     = {649--669},
  publisher = {Elsevier},
  title     = {Situation Calculus},
  year      = {2008},
  series    = {Foundations of Artificial Intelligence},
  volume    = {3},
  booktitle = {Handbook of Knowledge Representation},
  doi       = {10.1016/s1574-6526(07)03016-7},
}

@InProceedings{LopezSanchez2013,
  author    = {López-Sánchez, Javier and Rodríguez-Aguilar, Maite and Wooldridge, Juan and Vasconcelos, Michael and null, Wamberto},
  booktitle = {Proceedings of the 12th International Conference onAutonomous Agents and Multiagent Systems (AAMAS 2013)},
  title     = {Automated Synthesis of Normative Systems},
  year      = {2013},
}

@Article{Onn1997,
  author    = {Shmuel Onn and Moshe Tennenholtz},
  journal   = {Artificial Intelligence},
  title     = {Determination of social laws for multi-agent mobilization},
  year      = {1997},
  number    = {1},
  pages     = {155--167},
  volume    = {95},
  doi       = {10.1016/s0004-3702(97)00045-3},
  publisher = {Elsevier {BV}},
}

@Article{Crawford1995,
  author    = {Sue E. S. Crawford and Elinor Ostrom},
  journal   = {American Political Science Review},
  title     = {A Grammar of Institutions},
  year      = {1995},
  number    = {3},
  pages     = {582--600},
  volume    = {89},
  doi       = {10.2307/2082975},
  publisher = {Cambridge University Press ({CUP})},
}

@Article{Schiffel2014,
  author    = {S. Schiffel and M. Thielscher},
  journal   = {Journal of Artificial Intelligence Research},
  title     = {Representing and Reasoning About the Rules of General Games With Imperfect Information},
  year      = {2014},
  pages     = {171--206},
  volume    = {49},
  doi       = {10.1613/jair.4115},
  publisher = {{AI} Access Foundation},
}

@Article{Shoham1995,
  author    = {Yoav Shoham and Moshe Tennenholtz},
  journal   = {Artificial Intelligence},
  title     = {On social laws for artificial agent societies: off-line design},
  year      = {1995},
  number    = {1-2},
  pages     = {231--252},
  volume    = {73},
  doi       = {10.1016/0004-3702(94)00007-n},
  publisher = {Elsevier {BV}},
}

@Article{Smajgl2008,
  author    = {Alex Smajgl and Luis R. Izquierdo and MArco Huigne},
  journal   = {Advances in Complex Systems},
  title     = {Modeling Endogenous Rule Changes in an Institutional Context: the ADICO Sequence},
  year      = {2008},
  number    = {02},
  pages     = {199--215},
  volume    = {11},
  doi       = {10.1142/s021952590800157x},
  publisher = {World Scientific Pub Co Pte Lt},
}

@Article{Genesereth2005,
  author  = {Genesereth, Michael and Love, Nathaniel and Pell, Barney},
  journal = {AI Magazine},
  title   = {General Game Playing: Overview of the AAAI Competition.},
  year    = {2005},
  month   = {06},
  pages   = {62-72},
  volume  = {26},
  doi     = {10.1609/aimag.v26i2.1813},
}

@Article{Ostrom2011,
  author    = {Elinor Ostrom},
  journal   = {Policy Studies Journal},
  title     = {Background on the Institutional Analysis and Development Framework},
  year      = {2011},
  number    = {1},
  pages     = {7--27},
  volume    = {39},
  doi       = {10.1111/j.1541-0072.2010.00394.x},
  publisher = {Wiley},
}

@Book{Ostrom2005,
  author    = {Ostrom, Elinor},
  publisher = {Princeton University Press},
  title     = {Understanding Institutional Diversity},
  year      = {2005},
  isbn      = {0691122385},
  month     = sep,
  ean       = {9780691122380},
  pagetotal = {376},
}

@Book{Black1962,
  author    = {Black, Max},
  publisher = {Cornell University Press},
  title     = {Models and Metaphors: Studies in Language and Philosophy},
  year      = {1962},
  address   = {Ithaca, NY},
  isbn      = {9781501741326},
}

@InBook{Weymark2016,
  author    = {John Weymark},
  editor    = {Matthew D. Adler and Marc Fleurbaey},
  publisher = {Oxford University Press},
  title     = {Social Welfare Functions},
  year      = {2016},
  booktitle = {The Oxford Handbook of Well-Being and Public Policy},
  doi       = {10.1093/oxfordhb/9780199325818.013.5},
}

@Article{Sarr2021,
  author  = {Sarr, S. and Hayes, B. and DeCaro, D. A.},
  journal = {Land Use Policy},
  title   = {Applying Ostrom's Institutional Analysis and Development Framework, and Design Principles for Co-Production to Pollution Management in Louisville's Rubbertown, Kentucky},
  year    = {2021},
  month   = may,
  pages   = {105383},
  volume  = {104},
  af      = {Sarr, SaitEOLEOLHayes, BunnyEOLEOLDeCaro, Daniel A.},
  doi     = {10.1016/j.lusepol.2021.105383},
  ei      = {1873-5754},
  sn      = {0264-8377},
  tc      = {0},
  ut      = {WOS:000634547600007},
  z9      = {0},
}

@Article{Nguyen2020,
  author  = {Nguyen, T. and Watanabe, T.},
  journal = {Sustainability},
  title   = {Autonomous Motivation for the Successful Implementation of Waste Management Policy: An Examination Using an Adapted Institutional Analysis and Development Framework in Thua Thien Hue, Vietnam},
  year    = {2020},
  month   = apr,
  number  = {7},
  pages   = {2724},
  volume  = {12},
  af      = {Tam NguyenEOLEOLWatanabe, Tsunemi},
  doi     = {10.3390/su12072724},
  ei      = {2071-1050},
  tc      = {0},
  ut      = {WOS:000531558100149},
  z9      = {0},
}

@Article{Barton2017,
  author  = {Barton, D. N. and Benavides, K. and Chacon-Cascante, A. and Le Coq, J. F. and Quiros, M. M. and Porras, I. and Primmer, E. and Ring, I.},
  journal = {Environmental Policy and Governance},
  title   = {Payments for Ecosystem Services as a Policy Mix: Demonstrating the Institutional Analysis and Development Framework on Conservation Policy Instruments},
  year    = {2017},
  number  = {5},
  pages   = {404--421},
  volume  = {27},
  af      = {Barton, David N.EOLEOLBenavides, KarlaEOLEOLChacon-Cascante, AdrianaEOLEOLLe Coq, Jean-FrancoisEOLEOLQuiros, Miriam MirandaEOLEOLPorras, InaEOLEOLPrimmer, EevaEOLEOLRing, Irene},
  doi     = {10.1002/eet.1769},
  ei      = {1756-9338},
  oi      = {Chacon-Cascante, Adriana/0000-0002-9326-4659; Primmer,EOLEOLEeva/0000-0001-8954-8205; Ring, Irene/0000-0002-2688-8947},
  ri      = {Xu, Tianying/AAB-8185-2020; Chacon-Cascante, Adriana/AAH-2140-2019;EOLEOLBITOUN, RACHEL/AAC-9538-2021},
  si      = {SI},
  sn      = {1756-932X},
  tc      = {19},
  ut      = {WOS:000413350500002},
  z9      = {20},
}

@InBook{Kiser1982,
  author    = {Larry L. Kiser and Elinor Ostrom},
  chapter   = {2},
  editor    = {Michael D. McGinnis},
  pages     = {56-88},
  publisher = {Michigan University Press},
  title     = {The Three Worlds of Action: A Metatheoretical Synthesis of Institutional Approaches},
  year      = {1982},
  address   = {Ann Arbor},
}

@Article{Frantz2021,
  author    = {Christopher K. Frantz and Saba Siddiki},
  journal   = {Public Administration},
  title     = {Institutional Grammar 2.0: A specification for encoding and analyzing institutional design},
  year      = {2021},
  doi       = {10.1111/padm.12719},
  publisher = {Wiley},
}

@Article{Thielscher2016,
  author    = {Michael Thielscher},
  journal   = {Frontiers in Artificial Intelligence and Applications},
  title     = {GDL-III: A Proposal to Extend the Game Description Language to General Epistemic Games},
  year      = {2016},
  issn      = {0922-6389},
  pages     = {1630-1631},
  volume    = {285},
  doi       = {10.3233/978-1-61499-672-9-1630},
  publisher = {IOS Press},
}

@Article{Jonge2017,
  author    = {Dave de Jonge and Tomas Trescak and Carles Sierra and Simeon Simoff and Ramon L{\'{o}}pez de M{\'{a}}ntaras},
  journal   = {{AI} {\&} {SOCIETY}},
  title     = {Using Game Description Language for mediated dispute resolution},
  year      = {2017},
  number    = {4},
  pages     = {767--784},
  volume    = {34},
  doi       = {10.1007/s00146-017-0790-8},
  publisher = {Springer Science and Business Media {LLC}},
}

@Article{Jonge2021,
  author    = {Dave de Jonge and Dongmo Zhang},
  journal   = {Autonomous Agents and Multi-Agent Systems},
  title     = {{GDL} as a unifying domain description language for declarative automated negotiation},
  year      = {2021},
  number    = {1},
  volume    = {35},
  doi       = {10.1007/s10458-020-09491-6},
  publisher = {Springer Science and Business Media {LLC}},
}

@InCollection{Grossi2010,
  author    = {D. Grossi and D. Gabbay and L. van der Torre},
  booktitle = {Specification and Verification of Multi-agent Systems},
  publisher = {Springer {US}},
  title     = {The Norm Implementation Problem in Normative Multi-Agent Systems},
  year      = {2010},
  pages     = {195--224},
  doi       = {10.1007/978-1-4419-6984-2_7},
}

@Article{Wright1951,
  author    = {G. H. von Wright},
  journal   = {Mind},
  title     = {Deontic Logic},
  year      = {1951},
  issn      = {00264423, 14602113},
  number    = {237},
  pages     = {1--15},
  volume    = {60},
  publisher = {[Oxford University Press, Mind Association]},
  url       = {http://www.jstor.org/stable/2251395},
}

@InCollection{Kuhn1953,
  author    = {H. W. Kuhn},
  booktitle = {Contributions to the Theory of Games ({AM}-28), Volume {II}},
  publisher = {Princeton University Press},
  title     = {11. Extensive Games and the Problem of Information},
  year      = {1953},
  pages     = {193--216},
  doi       = {10.1515/9781400881970-012},
}

@Book{Ostrom1994,
  author    = {Elinor Ostrom and Roy Gardner and Jimmy Walker},
  publisher = {University of Michigan Press},
  title     = {Rules, Games, and Common-Pool Resources},
  year      = {1994},
  doi       = {10.3998/mpub.9739},
}

@Article{Andrighetto2012,
  author    = {Giulia Andrighetto and Guido Governatori and Pablo Noriega and Leon van der Torre},
  journal   = {Dagstuhl Reports},
  title     = {{Normative Multi-Agent Systems (Dagstuhl Seminar 12111)}},
  year      = {2012},
  issn      = {2192-5283},
  number    = {3},
  pages     = {23--49},
  volume    = {2},
  address   = {Dagstuhl, Germany},
  doi       = {10.4230/DagRep.2.3.23},
  editor    = {Giulia Andrighetto and Guido Governatori and Pablo Noriega and Leon van der Torre},
  publisher = {Schloss Dagstuhl--Leibniz-Zentrum fuer Informatik},
  url       = {http://drops.dagstuhl.de/opus/volltexte/2012/3535},
  urn       = {urn:nbn:de:0030-drops-35358},
}

@InProceedings{Hahn2020,
  author    = {Carsten Hahn and Thomy Phan and Sebastian Feld and Christoph Roch and Fabian Ritz and Andreas Sedlmeier and Thomas Gabor and Claudia Linnhoff-Popien},
  booktitle = {Proceedings of the 12th International Conference on Agents and Artificial Intelligence},
  title     = {Nash Equilibria in Multi-Agent Swarms},
  year      = {2020},
  publisher = {{SCITEPRESS} - Science and Technology Publications},
  doi       = {10.5220/0008990802340241},
}

@Book{Shoham2014,
  author    = {Yoav Shoham and Kevin Leyton-Brown},
  publisher = {Cambridge University Press},
  title     = {Multiagent Systems: Algorithmic, Game-Theoretic, and Logical Foundations},
  year      = {2014},
  isbn      = {0521899435},
  month     = oct,
  ean       = {9780521899437},
  pagetotal = {504},
}

@Article{Caillou2009,
  author    = {Philippe Caillou and Samir Aknine and Suzanne Pinson},
  journal   = {Group Decision and Negotiation},
  title     = {Searching Pareto Optimal Solutions for the Problem of Forming and Restructuring Coalitions in Multi-Agent Systems},
  year      = {2009},
  number    = {1},
  pages     = {7--37},
  volume    = {19},
  doi       = {10.1007/s10726-009-9183-9},
  publisher = {Springer Science and Business Media {LLC}},
}

\newpage

\appendix
\section{Algorithms}\label{sec:algorithms}
For every function pseudo-code, the following is specified:
\begin{itemize}
	\item \textbf{Input:} its arguments.
	\item \textbf{Output:} its return value(s) and/or data structure(s).
	\item \textbf{Data:} the information stored in the database being consulted. In general, all functions call upon the original action situation description $\mathbb{A}$. Additionally, the facts corresponding to the current state $s_t$ and/or the action profile executed $\mu$ might also be necessary.
	\item \textbf{Description:} Short documentation on the procedure implemented by the function.
\end{itemize}

\begin{algorithm}
	\KwInput{\textit{id} $\triangleright$ string\newline
		\textit{type} $\triangleright$ one of either ``boundary'', ``position'' or ``choice''\newline
		\textit{thres} $\triangleright$ non-negative integer}
	\KwOutput{$\mathcal{C}$ $\triangleright$ a set of ground atoms}
	\KwData{$\mathbb{A}$ $\triangleright$ an action situation description\newline
		$\phi = \{\texttt{participates}(ag_i)\}$ $\triangleright$ set of participant fluents (if \textit{type}=``position'')\newline
		$\rho = \{\texttt{role}(ag_i, r_i)\}$ $\triangleright$ set of role fluents (if \textit{type}=``choice'')\newline
		$s_t = \{f_1,...,f_n\}$ $\triangleright$ set of state fluents (if \textit{type}=``choice'')}
	\KwDescription{Get the participants, roles or available actions entailed by the boundary, position or choice rules respectively.}
	
	\SetKwFunction{GetSimpleConseqs}{Get-Simple-Conseqs}
	\SetKwProg{Fn}{Function}{:}{}
	
	\Fn{\GetSimpleConseqs{\textit{id}, \textit{type}, \textit{thres}}}{
		$kv \leftarrow []$\;
		\ForEach{instantiation of \mintinline{prolog}{?-query_rule(rule(}\textit{id},\textit{type},$n$,\mintinline{prolog}{if Cond then} $f$ \mintinline{prolog}{where Constr))}}{
			\lIf{$n \leq \textit{thres}$}{$kv$.\textsc{Append}($n$ : $f$)}
		}
		$kv \leftarrow \textsc{Sort-By-Descending-Key}(kv)$\;
		$\mathcal{C} \leftarrow \{\}$\;
		\For{($n$ : $f$) pair in $kv$}{
			\lIf{\mintinline{prolog}{Conseq} $\notin \mathcal{C}$ \And \mintinline{prolog}{~Conseq} $\notin \mathcal{C}$}{$\mathcal{C} \leftarrow \mathcal{C} \cup \{\mintinline{prolog}{Conseq}\}$}
		}
		$\mathcal{C} \leftarrow \{c \in \mathcal{C} \mid c \neq \sim f\}$\;
		\Return{$\mathcal{C}$}
	}
	\caption{Function \textsc{Get-Simple-Conseqs}(\textit{id}, \textit{type}, \textit{thres})}
	\label{alg:get-simple-conseqs}
\end{algorithm}

\begin{algorithm}
	\KwInput{\textit{id} $\triangleright$ string\newline
		\textit{thres} $\triangleright$ non-negative integer}
	\KwOutput{$\mathcal{S}_{t+1} = \{s_{t+1}^1,s_{t+1}^2,...\}$ $\triangleright$ the set of potential next states, each corresponds to a set of fluents\newline
		$\mathcal{P}: \mathcal{S}_{t+1} \rightarrow [0,1]$ $\triangleright$ a probability distribution over the next states}
	\KwData{$\mathbb{A}$ $\triangleright$ action situation description\newline
		$s_t = \{\mathit{f_1},\mathit{f_2},...\}$ $\triangleright$ set of facts that hold true at the current state\newline
		$\mu = \{\texttt{does}(ag_1,ac_1), \texttt{does}(ag_2,ac_2), ...\}$ $\triangleright$ joint action profile}
	\KwDescription{Get the post-transition state fluents and their probabilities that derive from performing some joint action in a pre-transition state.}
	
	\SetKwFunction{GetControlConseqs}{Get-Control-Conseqs}
	\SetKwProg{Fn}{Function}{:}{}
	
	\Fn{\GetControlConseqs{\textit{id}, \textit{thres}}}{
		$kv \leftarrow []$\;
		\ForEach{instantiation of \mintinline{prolog}{?-query_rule(rule(}\textit{id},\mintinline{prolog}{control},$n$,\mintinline{prolog}{if Cond then} $conseqs$ \mintinline{prolog}{where Constr))}}{
			\lIf{$n \leq \textit{thres}$}{$kv$.\textsc{Append}($n$ : $conseqs$)}
		}
		$kv \leftarrow \textsc{Sort-By-Descending-Key}(kv)$\;
		$\mathcal{S}_{t+1} \leftarrow \{\{\}\}$\;
		$\mathcal{P}(\{\}) = 1$\;
		
		\For(\tcp*[f]{loop over activated control rules}){($n$ : $conseqs$) pair in $kv$}{
			$conseq s= [c_{11}$ \mintinline{prolog}{and} $c_{12}$ \mintinline{prolog}{and} ... \mintinline{prolog}{withProb} $p_1$, \newline
			\hspace*{4.7em}$c_{21}$ \mintinline{prolog}{and} $c_{22}$ \mintinline{prolog}{and} ... \mintinline{prolog}{withProb} $p_2$,\newline
			\hspace*{4.7em} ... $]$\;
			
			\tcc{check that every rule is consistent with the facts already established in the potential next states}
			\For{($c_{i1}$  \mintinline{prolog}{and} $c_{i2}$ \mintinline{prolog}{and} ... \mintinline{prolog}{withProb} $p_i$) in $conseqs$}{
				$\mathcal{C}_i \leftarrow \{c_{ij}\}_{c_i = c_{i1} \; \texttt{and} \; c_{i2} \; \texttt{and} \; ...}$\;
				\For{$(c_{ij},s_{t+1})$ in $\mathcal{C}_i \times \mathcal{S}_{t+1}$}{
					\lIf{\mintinline{prolog}{?-incompatible}($c_{ij}$, $s_{t+1}$) returns \mintinline{prolog}{true}}{\GoTo line 9}
				}
			}
			
			\tcc{the activated rule consequences are consistent with $\mathcal{S}_{t+1}$}
			$\mathcal{S}'_{t+1} \leftarrow \{\}$\;
			\For{$s_{t+1} \in \mathcal{S}_{t+1}$}{
				\For{($c_{i1}$  \mintinline{prolog}{and} $c_{i2}$ \mintinline{prolog}{and} ... \mintinline{prolog}{withProb} $p_i$) in $conseqs$}{
					$\mathcal{C}_i \leftarrow \{c_{ij}\}_{c_i = c_{i1} \; \texttt{and} \; c_{i2} \; \texttt{and} \; ...}$\;
					$\mathcal{S}'_{t+1} \leftarrow \mathcal{S}'_{t+1} \cup \{s_{t+1} \cup \mathcal{C}_i\}$\;
					$\mathcal{P}(s_{t+1} \cup \mathcal{C}_i) \leftarrow \mathcal{P}(s_{t+1}) \cdot p_i$\;
				}
			}
			$\mathcal{S}_{t+1} \leftarrow \mathcal{S}'_{t+1}$\;
		}
		
		\For(\tcp*[f]{drag compatible facts from $s_t$ over to $s_{t+1}$}){$(\mathit{f_i},s_{t+1}) \in s_t \times \mathcal{S}_{t+1}$}{
			\lIf{\mintinline{prolog}{?-incompatible}($\mathit{f_i}$, $s_{t+1}$) returns \mintinline{prolog}{false}}{$s_{t+1} \leftarrow s_{t+1} \cup \{\mathit{f_i}\}$}
		}
		\Return{$\mathcal{S}_{t+1}$, $\mathcal{P}$}
	}
	\caption{Function \textsc{Get-Control-Conseqs}(\textit{id}, \textit{thres})}
	\label{alg:get-control-conseqs}
\end{algorithm}

\begin{algorithm}
	\KwInput{\textit{id} $\triangleright$ string\newline
		\textit{thres} $\triangleright$ non-negative integer}
	\KwOutput{$\gamma$ $\triangleright$ game round\newline
		$F$ $\triangleright$ set of fluents assigned to $\gamma$'s terminal nodes\newline
		$\tau: Z \rightarrow \{0,1\}$ $\triangleright$ termination conditions are met at $\gamma$'s terminal nodes}
	\KwData{$\mathbb{A}$ $\triangleright$ action situation description\newline
		$s_t = \{\mathit{f_1}, \mathit{f_2}, ...\}$ $\triangleright$ set of facts}
	\KwDescription{Given a state characterized by a set of facts, model all the ways by which it may evolve as a game round.}
	
	\SetKwFunction{BuildGameRound}{Build-Game-Round}
	\SetKwProg{Fn}{Function}{:}{}

	\Fn{\BuildGameRound{\textit{id}, \textit{thres}}}{
		$k \leftarrow 1$, $X \leftarrow \{k\}$, $x_0 \leftarrow k$, $k++$\;
		$E \leftarrow \{\}$\;
		
		\tcc{set players to those participants that can take some action}
		$M \leftarrow \textsc{Get-Simple-Conseqs}$(\textit{id}, choice, \textit{thres}) = $\{\texttt{can}(ag_1,ac_1), \texttt{can}(ag_2, ac_2), ...\}$\;
		$P \leftarrow \{ag_i \mid \exists ac, \texttt{can}(ag_i, ac) \in M\}$\;
		
		\tcc{STEP 1: Build the game tree in a breadth-first manner}
		$w \leftarrow \{x_0\}$, $w' \leftarrow \{\}$ \tcp*{current and next information sets}
		$W \leftarrow \{\}$, $\mathscr{A} \leftarrow \{\}$\tcp*{information partition and actions}
		\For{$\mathit{player} \in P$}{
			$W_{\mathit{player}} \leftarrow \{\mathit{w}\}$, $W \leftarrow W \cup \{W_{\mathit{player}}\}$\;
			$A(w) \leftarrow \{ac \mid \texttt{can}(\mathit{player}, ac) \in M\}$, $\mathscr{A} \leftarrow \mathscr{A} \cup \{A(w)\}$\;
			\For{$x \in w$}{
				$T(x) \leftarrow \mathit{player}$\;
				\For{$\mathit{action} \in A(w)$}{
					$X \leftarrow X \cup \{k\}$, $w' \leftarrow w' \cup \{k\}$\;
					$E \leftarrow E \cup \{(x,k)\}$, $\mathit{label}(x,k) \leftarrow \mathit{action}$\;
					$k++$\;
				}
			}
			$w \leftarrow w'$, $w' \leftarrow \{\}$\;
		}
		
		\tcc{STEP 2: Get the facts and chance moves at the terminal nodes}
		$p \leftarrow \{\}$\tcp*{probability over chance moves}
		\For(\tcp*[f]{$Z$ is the subset of terminal nodes}){$z \in Z \subseteq X$}{
			\tcp{action profile from root to terminal node}
			$\mu = \{\texttt{does}(\mathit{ag}, \mathit{ac})\}_{\forall \mathit{ag} = T(x_i), \; \mathit{ac} = \mathit{label}(x_i, x_{i+1}) \; \mid \; (x_i, x_{i+1}) \in \textsc{Path}(x_0, z)}$\;
			$\mathbb{A} \leftarrow \mathbb{A} \cup \mu$ \tcp*{assert action profile into database}
			\leIf{\mintinline{prolog}{?-terminal} returns \mintinline{prolog}{true}}{$t \leftarrow 1$}{$t \leftarrow 0$}
			$\mathcal{S}_{t+1}, \mathcal{P} = \textsc{Get-Control-Conseqs}$(\textit{id}, \textit{thres})\;
			\lIf{$\mathcal{S}_{t+1} = \{s_{t+1}\}$}{$F(z) = s_{t+1}$, $\tau(z) = t$ \tcp*[f]{no stochastic effects}}
			\Else(\tcp*[f]{stochastic effects}){
				$T(z) \leftarrow \mathit{chance}$\;
				\For{$s_{t+1}^i \in \mathcal{S}_{t+1}$}{
					$X \leftarrow X \cup \{k\}$, $E \leftarrow E \cup \{(z,k)\}$\;
					$p_z(z,k) \leftarrow \mathcal{P}\left(s_{t+1}^i\right)$\;
					$F(k) \leftarrow s_{t+1}^i$, $\tau(k) \leftarrow t$, $k++$\;
				}
				$p \leftarrow p \cup \{p_z\}$\;
			}
			$\mathbb{A} \leftarrow \mathbb{A} \setminus \mu$\tcp*{remove actions from the database}
		}
		
		$\gamma = \left(P, (X,E), T, W, \mathscr{A}, p\right)$\;
		\Return{$\gamma$, $F$, $\tau$}
	}
	\caption{Function \textsc{Build-Game-Round}(\textit{id}, \textit{thres})}
	\label{alg:build-game-round}
\end{algorithm}

\begin{algorithm}
	\KwInput{\textit{id} $\triangleright$ string\newline
		\textit{thres} $\triangleright$ non-negative integer\newline
		\textit{max} $\triangleright$ non-negative integer}
	\KwOutput{$\Gamma$ $\triangleright$ extensive-form game\newline
		$\mathcal{F}$ $\triangleright$ set of fluents assigned to $\gamma$'s state nodes}
	\KwData{$\mathbb{A}$ $\triangleright$ action situation description}
	\KwDescription{Given an action situation description, generate its extensive-form game semantics.}
	
	\SetKwFunction{BuildFullGame}{Build-Full-Game}
	\SetKwProg{Fn}{Function}{:}{}

	\Fn{\BuildGameRound{\textit{id}, \textit{thres}, \textit{max}}}{
		$\phi \leftarrow \textsc{Get-Simple-Conseqs}(\textit{id}, \text{boundary}, \textit{thres}) = \{\texttt{participates}(ag_1), ...\} $\;
		$\mathbb{A} \leftarrow \mathbb{A} \cup \phi$\;
		$\rho \leftarrow \textsc{Get-Simple-Conseqs}(\textit{id}, \text{position}, \textit{thres}) = \{\texttt{role}(ag_1,r_1), ...\} $\;
		$\mathbb{A} \leftarrow \mathbb{A} \cup \rho$\;
		$s_0 \leftarrow \{\}$\;
		\lForEach(\tcp*[f]{initial facts}){instantiation $f_i$ of \mintinline{prolog}{?-initially(F)}}{$s_0 \leftarrow s_0 \cup \{f_i\}$}
		$P=\{ag_i\}_{\forall \texttt{participates}(ag_i) \in \phi}$\;
		$X \leftarrow \{1\}$, $x_0 \leftarrow 1$, $E \leftarrow \{\}$, $W \leftarrow \{\{\},...,\{\}\}_{\forall i \in P}$, $\mathscr{A} \leftarrow \{\}$, $p \leftarrow \{\}$\;
		$\mathcal{F}(1) \leftarrow s_0$\;
		$\textit{round}(1) \leftarrow 0$\;
		$\mathcal{Q} \leftarrow \textsc{Queue}(1)$\;
		
		\While{$\mathcal{Q}$ not empty}{
			$n \leftarrow \mathcal{Q}.\textsc{Pop}()$\;
			\lIf{$\textit{round}(n) \geq \textit{max}$}{\Continue}
			$s_t \leftarrow \mathcal{F}(n)$\;
			$\mathbb{A} \leftarrow \mathbb{A} \cup \{s_t\}´$\tcp*{assert node facts into database}
			\lIf{\mintinline{prolog}{?-terminal} returns \mintinline{prolog}{true}}{$\mathbb{A} \leftarrow \mathbb{A} \setminus \{s_t\}$, \Continue}
			$\gamma, F, \tau \leftarrow \textsc{Build-Game-Round}$(\textit{id}, \textit{thres}) \;
			$\mathbb{A} \leftarrow \mathbb{A} \setminus \{s_t\}$\;
			\tcc{append game round to overall game tree -- superindex $\gamma$ denotes the elements from the game round tuple}
			\lFor(\tcp*[f]{node re-labeling}){$x \in X^{\gamma}$}{$x \leftarrow x+n-1$}
			$X \leftarrow X \cup X^{\gamma}$, $E \leftarrow E \cup E^{\gamma}$\;
			\lFor{$x \in X^{\gamma} \setminus Z^{\gamma}$}{$T(x) \leftarrow T^{\gamma}(x)$}
			\lFor{$p \in P$}{$W_p \leftarrow W_p \cup W_p^{\gamma}$}
			\lFor{$A(w) \in \mathscr{A}^{\gamma}$}{$\mathscr{A} \leftarrow \mathscr{A} \cup \{A(w)\}$}
			\lForAll{$x \in X^{\gamma} \mid T(x) = \textit{chance}$}{$p \leftarrow p \cup \{p_x^{\gamma}\}$}
			\lFor{$z \in Z^{\gamma}$}{$\mathcal{F}(z) \leftarrow F(z)$}
			\ForAll{$z \in Z^{\gamma}$}{
				$\textit{round}(z) \leftarrow \textit{round}(n) + 1$\;
				\lIf{ $\tau(z) = 0$}{$\mathcal{Q}.\textsc{Push}(z)$}
			}
		}
		$\Gamma = \left(P, (X,E), T, W, \mathscr{A}, p\right)$\;
		\Return{$\Gamma$, $\mathcal{F}$}
	}
	\caption{Function \textsc{Build-Full-Game}(\textit{id}, \textit{thres}, \textit{max})}
	\label{alg:build-full-game}
\end{algorithm}

\newpage

\section{Iterated Prisoner's Dilemma example game trees}\label{sec:ipd-game-trees}
\begin{sidewaysfigure}[h]
	\centering
	\includegraphics[width=\linewidth]{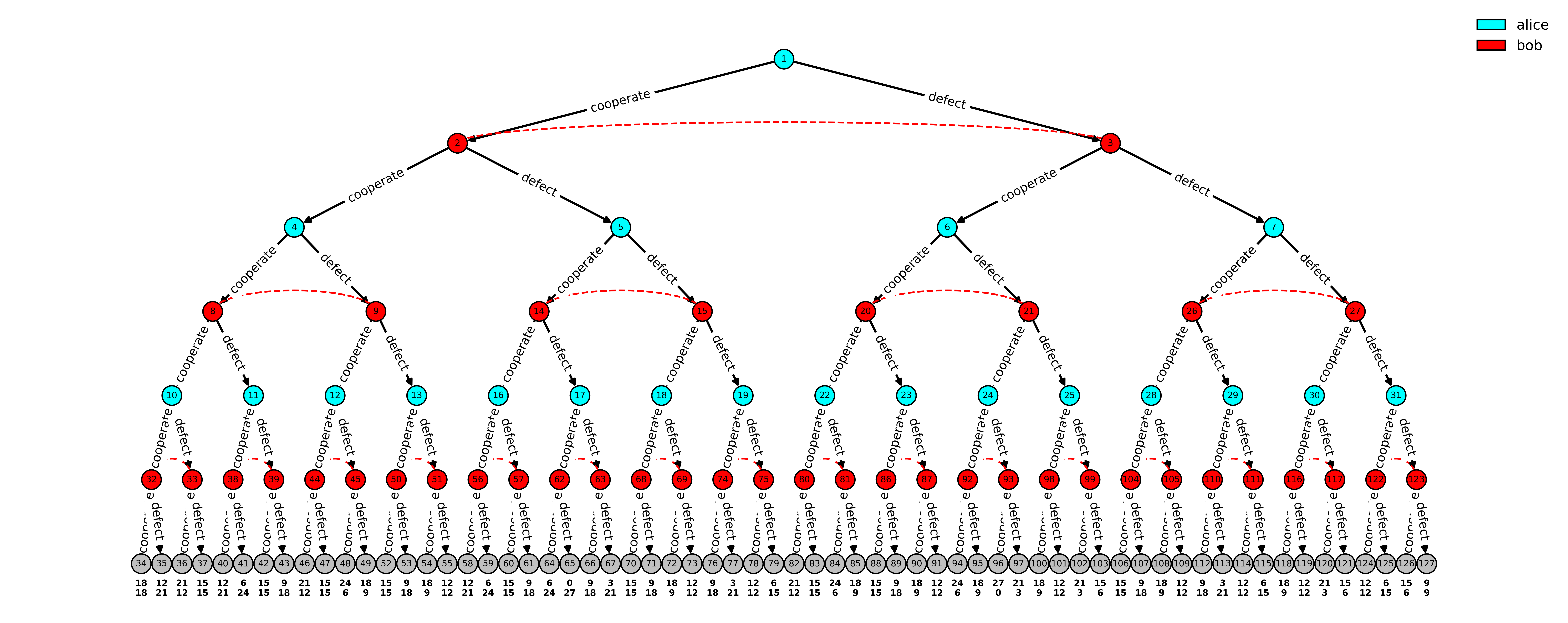}
	\caption{Game semantics for the iterated Prisoner's Dilemma with the default rule base.}
	\label{fig:ipd-default}
\end{sidewaysfigure}
\newpage
\begin{sidewaysfigure}
	\centering
	\includegraphics[width=\linewidth]{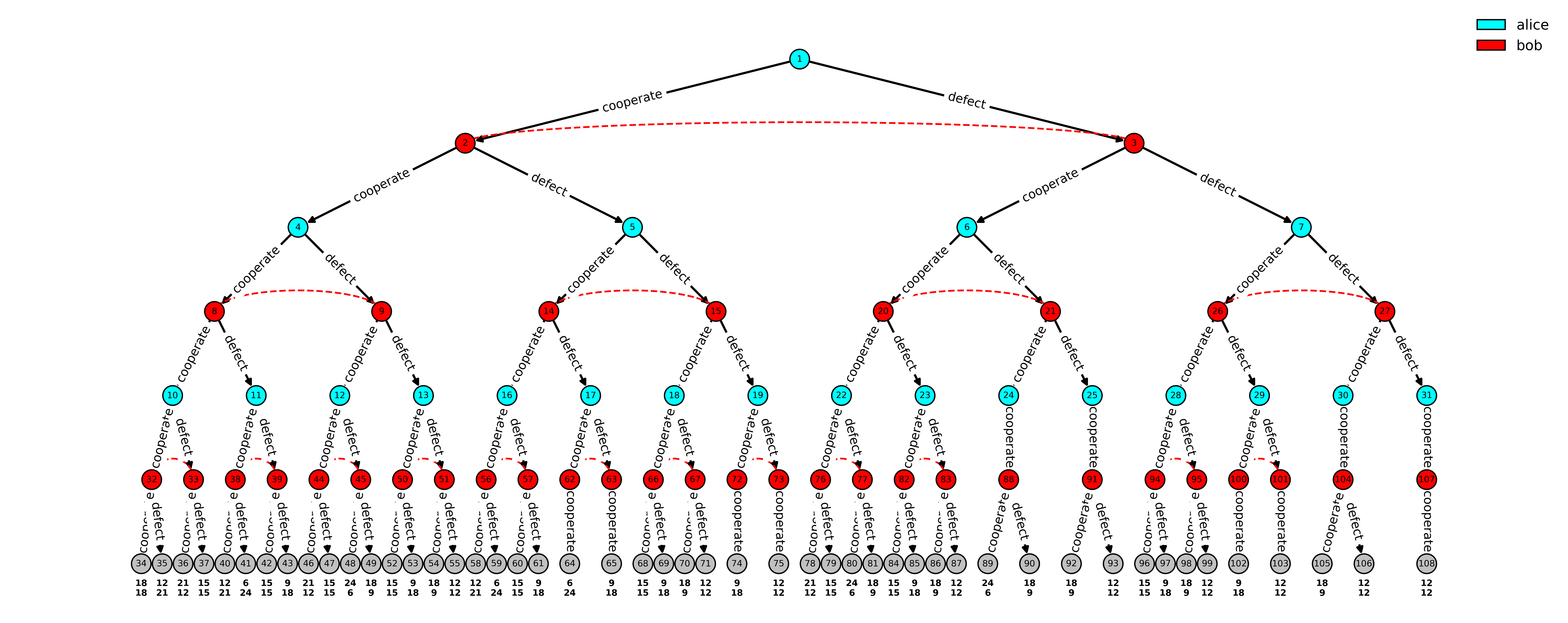}
	\caption{Game semantics for the iterated Prisoner's Dilemma with an additional rule for restricting the number of consecutive defections.}
	\label{fig:ipd-limit-defections}
\end{sidewaysfigure}
\newpage
\begin{sidewaysfigure}
	\centering
	\includegraphics[width=\linewidth]{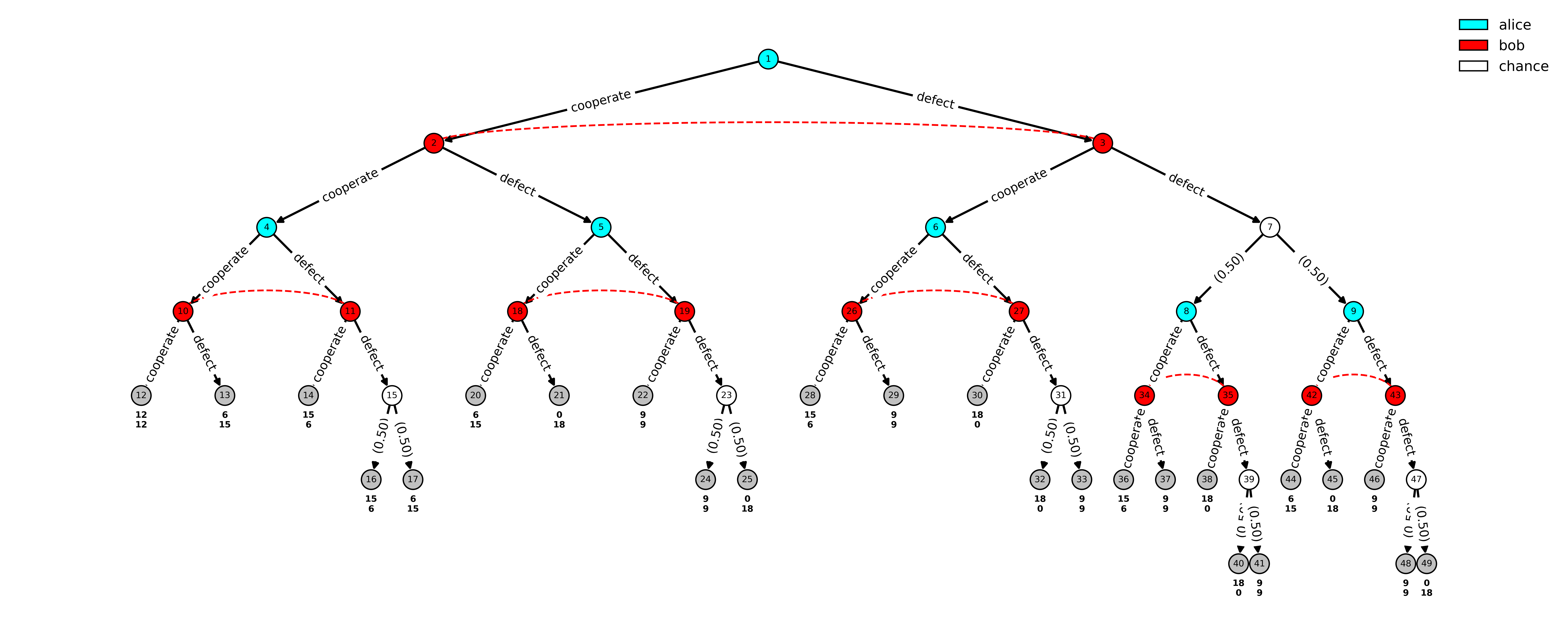}
	\caption{Game semantics for the iterated Prisoner's Dilemma with additional rules to change the outcome of a consecutive defection. For visualization purposes, termination conditions are met after agents play two game rounds instead of three.}
	\label{fig:ipd-ban-mutual-defect}
\end{sidewaysfigure}

\clearpage

\section{Example ASL descriptions}\label{sec:example-ASL-descriptions}
\begin{listing}[h!]
\begin{minipage}{0.2\textwidth}
\begin{minted}{prolog}
agent(i).
agent(j).
agent(k).
\end{minted}
\end{minipage}
\hfill
\begin{minipage}{0.75\textwidth}
\begin{minted}{prolog}
initially(time(0)).
initially(payoff(P,0)) :- participates(P).

incompatible(seen(A,_),_) :- does(A,_).
incompatible(~seen(A,_),_) :- does(A,_).
incompatible(time(_),L) :- member(time(_),L).
incompatible(payoff(P,_),L) :- member(payoff(P,_),L).
\end{minted}
\end{minipage}
\begin{minted}{prolog}
/*** Default rules ***/
rule(metanorms,boundary,0,if agent(A) then participates(A) where []).
rule(metanorms,position,0,if participates(A) then role(A,individual) where [A=i]).
rule(metanorms,position,0,if participates(A) then role(A,monitor) where [A=j]).
rule(metanorms,choice,0,if role(A,individual) then can(A,defect) where [time(0)]).
rule(metanorms,choice,0,if role(A,individual) then can(A,~defect) where [time(0)]).
rule(metanorms,choice,0,if role(P2,monitor) then can(P2,sanction(P1)) where [seen(P2,P1)]).
rule(metanorms,choice,0,if role(P2,monitor) then can(P2,~sanction(P1)) where [seen(P2,P1)]).
rule(metanorms,control,0,if does(_,_) then [time(M) withProb 1] where [time(N),{M=N+1}]).
rule(metanorms,control,0,if does(P,defect) then [payoff(P,Y) withProb 1]
	where [payoff(P,X),{Y=X+3}]).
rule(metanorms,control,0,if does(P1,defect) then [payoff(P2,Y) withProb 1]
	where [payoff(P2,X),P2\=P1,{Y=X-1}]).
rule(metanorms,control,0,if does(P1,defect) then [seen(P2,P1) withProb 0.6,
	~seen(P2,P1) withProb 0.4] where [role(P1,individual),role(P2,monitor)]).
rule(metanorms,control,0,if does(P2,sanction(P1))
	then [payoff(P1,Y1) and payoff(P2,Y2) withProb 1]
	where [payoff(P1,X1),payoff(P2,X2),{Y1=X1-9},{Y2=X2-2}]).

/*** Introduce metamonitor ***/
rule(metanorms,position,1,if participates(A) then role(A,metamonitor) where [A=k]).
rule(metanorms,choice,1,if role(P2,metamonitor) then can(P2,sanction(P1))
	where [seen(P2,P1)]).
rule(metanorms,choice,1,if role(P2,metamonitor) then can(P2,~sanction(P1))
	where [seen(P2,P1)]).
rule(metanorms,control,1,if does(P1,~sanction(_))then [seen(P2,P1) withProb 0.6,
	~seen(P2,P1) withProb 0.4] where [role(P1,monitor),role(P2,metamonitor)]).
\end{minted}
\caption{ASL description for Axelrod's norms and metanorms games: agents (top left), states (top right) and rule base (bottom).}
\label{code:metanorms-description}
\end{listing}
\newpage
\begin{figure}[h!]
	\begin{subfigure}{0.5\textwidth}
		\centering
		\includegraphics[width=\textwidth]{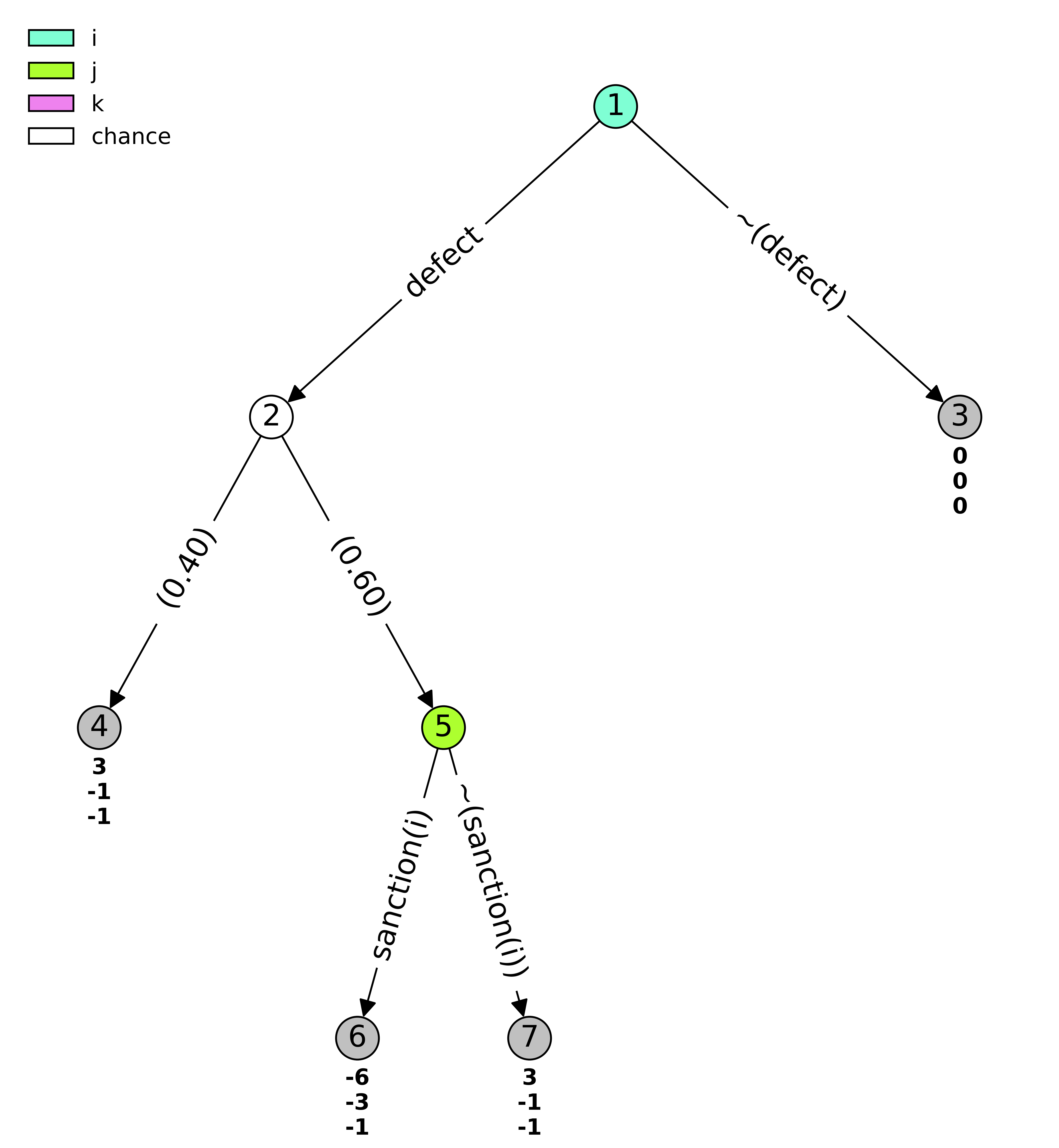}
		\caption{}
		\label{subfig:norms-game-tree}
	\end{subfigure}
	\begin{subfigure}{0.5\textwidth}
		\centering
		\small
		\begin{tabularx}{\textwidth}{|c|X|c|}
			\hline & \textbf{State fluents} & $\mathbf{p}$ \\
			\hline 1 & payoff(i,0), payoff(j,0), payoff(k,0), time(0) & - \\
			\hline 3 & payoff(i,0), payoff(j,0), payoff(k,0), time(1) & 0 \\
			\hline 4 & payoff(i,3), payoff(j,-1), payoff(k,-1), time(1), ~(seen(j,i)) & 0.4\\
			\hline 5 & payoff(i,3), payoff(j,-1), payoff(k,-1), seen(j,i), time(1) & - \\
			\hline 6 & payoff(i,-6), payoff(j,-3), payoff(k,-1), time(2) & 0 \\
			\hline 7 & payoff(i,3), payoff(j,-1), payoff(k,-1), time(2) & 0.6 \\
			\hline
		\end{tabularx}
		\caption{}
		\label{tab:norms-state-fluents}
	\end{subfigure}
	
	\begin{subfigure}{0.5\textwidth}
		\centering
		\includegraphics[width=\textwidth]{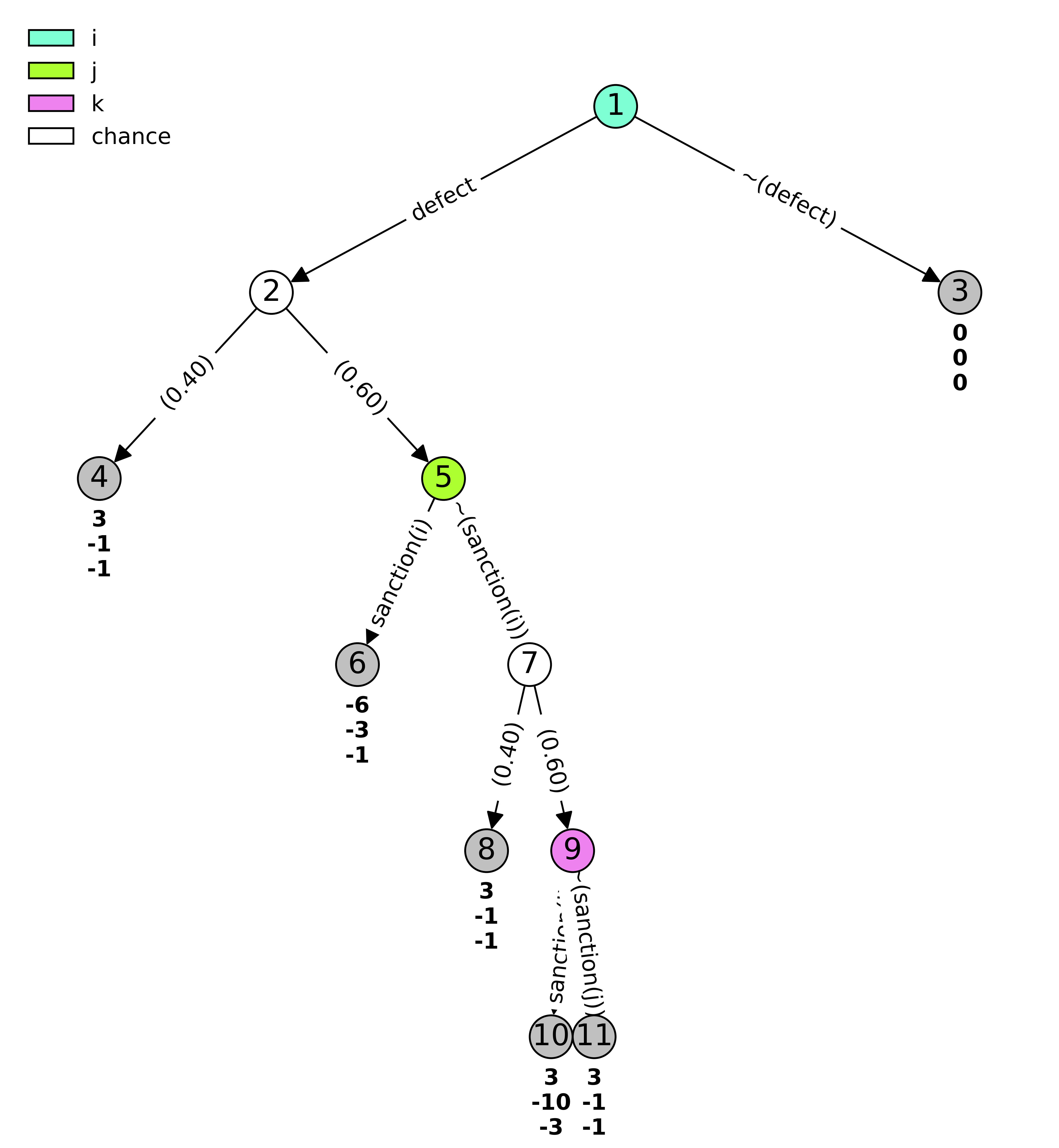}
		\caption{}
		\label{subfig:metanorms-game-tree}
	\end{subfigure}
	\begin{subfigure}{0.5\textwidth}
		\centering
		\small
		\begin{tabularx}{\textwidth}{|c|X|c|}
			\hline & \textbf{State fluents} & $\textbf{p}$ \\
			\hline 1 & payoff(i,0), payoff(j,0), payoff(k,0), time(0) & - \\
			\hline 3 & payoff(i,0), payoff(j,0), payoff(k,0), time(1) & 0\\
			\hline 4 & payoff(i,3), payoff(j,-1), payoff(k,-1), time(1), ~(seen(j,i)) & 0.4\\
			\hline 5 & payoff(i,3), payoff(j,-1), payoff(k,-1), seen(j,i), time(1) & - \\
			\hline 6 & payoff(i,-6), payoff(j,-3), payoff(k,-1), time(2) & 0 \\
			\hline 8 & payoff(i,3), payoff(j,-1), payoff(k,-1), time(2), ~(seen(k,j)) & 0.24 \\
			\hline 9 & payoff(i,3), payoff(j,-1), payoff(k,-1), seen(k,j), time(2) & - \\
			\hline 10 & payoff(i,3), payoff(j,-10), payoff(k,-3), time(3) & 0 \\
			\hline 11 & payoff(i,3), payoff(j,-1), payoff(k,-1), time(3) & 0.36 \\
			\hline
		\end{tabularx}
		\caption{}
		\label{tab:metanorms-state-fluents}
	\end{subfigure}
	\caption{Semantics for Axelrod's norms (top) and metanorms (bottom) action situations, with the extensive game tree (left) and the corresponding state fluents (right). For the terminal nodes, their probability $\mathbf{p}$ is also given.}
	\label{fig:axelrod-semantics}
\end{figure}

\begin{listing}
\begin{minipage}{0.35\textwidth}
\begin{minted}{prolog}
agent(alice).
agent(bob).
strength(alice,5).
strength(bob,3).
speed(alice,5).
speed(bob,8).
\end{minted}
\end{minipage}
\hfill
\begin{minipage}{0.65\textwidth}
\begin{minted}{prolog}
fishing_spot(spot1). fishing_spot(spot2).

initially(at(F,shore)) :- role(F,fisher).

terminal :-
	at(F1,S1),at(F2,S2),fishing_spot(S1),fishing_spot(S2),
	F1\=F2,S1\=S2.
terminal :- lost_fight(_).

incompatible(at(F,_),L) :- member(at(F,_),L).
incompatible(lost_fight(_),L) :- member(lost_fight(_),L).
incompatible(lost_race(_),L) :- member(lost_race(_),L).
\end{minted}
\end{minipage}
\begin{minted}{prolog}
/*** Default rules***/
rule(fishers,boundary,0,if agent(A) then participates(A) where []).
rule(fishers,position,0,if participates(A) then role(A,fisher) where []).
rule(fishers,choice,0,if role(A,fisher) then can(A,go_to_spot(S))
	where [at(A,shore),fishing_spot(S)]).
rule(fishers,choice,0,if role(A,fisher) then can(A,stay) where [at(A,S),fishing_spot(S)]).
rule(fishers,choice,0,if role(A,fisher) then can(A,leave) where [at(A,S),fishing_spot(S)]).
rule(fishers,control,0,if does(A,go_to_spot(S)) then [at(A,S) withProb 1] where []).
rule(fishers,control,0,if does(A,leave) then [at(A,S2) withProb 1]
	where [at(A,S1),fishing_spot(S1),fishing_spot(S2),S1\=S2]).
rule(fishers,control,0,if does(F1,A) and does(F2,A) then [lost_fight(F1) withProb P1,
	lost_fight(F2) withProb P2] where [at(F1,S),at(F2,S),F1@<F2,fishing_spot(S),
	strength(F1,X1),strength(F2,X2),{P1=X1/(X1+X2)},{P2=X2/(X1+X2)}]).
\end{minted}
\caption{ASL description for Ostrom's fishing game (default rules only): agents (top left), states (top right) and rules (bottom).}
\label{code:fishing-game-description-default}
\end{listing}

\begin{figure}[h]
	\centering
	\begin{subfigure}[h]{\textwidth}
		\centering
		\includegraphics[width=\textwidth]{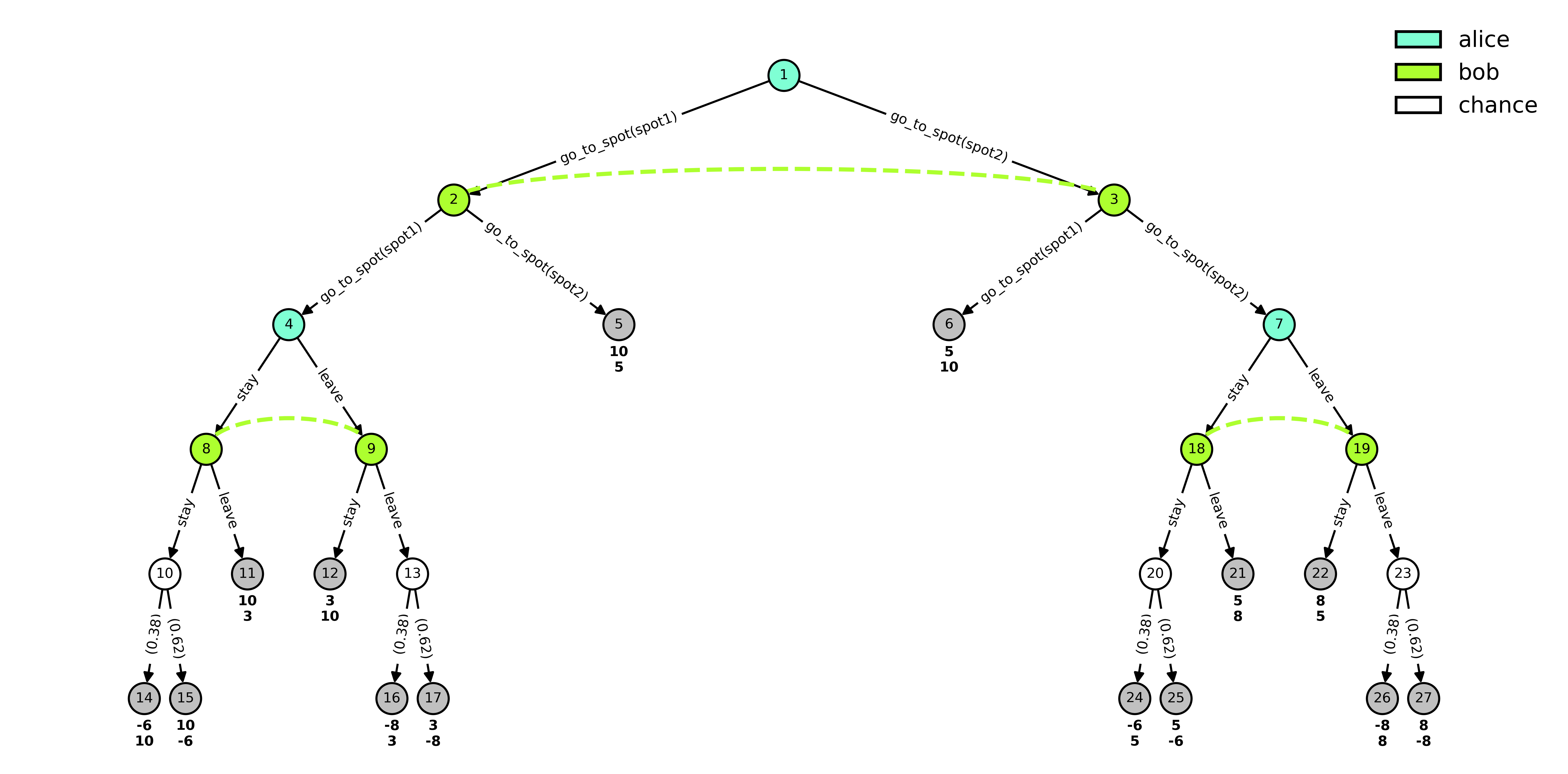}
		\caption{}
		\label{subfig:fishers-default-game-tree}
	\end{subfigure}

	\begin{subfigure}[h]{\textwidth}
		\centering
		\small
		\begin{tabular}{|c|l|c||c|l|c|}
			\hline & \textbf{State fluents} & $\mathbf{p}$ && \textbf{State fluents} & $\mathbf{p}$ \\
			\hline 1 & at(alice, shore), at(bob, shore) & - & 14 & at(alice, spot1), at(bob, spot1), lost\_fight(bob) & 0.18 \\
			\hline 4 & at(alice, spot1), at(bob, spot1) & - & 15 & at(alice, spot1), at(bob, spot1), lost\_fight(alice) & 0.31 \\
			\hline 5 & at(alice, spot1), at(bob, spot2) & 0.11 & 16 & at(alice, spot2), at(bob, spot2), lost\_fight(bob) & 0\\
			\hline 6 & at(alice, spot2), at(bob, spot1) & 0.25 & 17 & at(alice, spot2), at(bob, spot2), lost\_fight(alice) & 0 \\
			\hline 7 & at(alice, spot2), at(bob, spot2) &  - & 24 & at(alice, spot2), at(bob, spot2), lost\_fight(bob) & 0 \\
			\hline 11 & at(alice, spot1), at(bob, spot2) & 0 & 25 & at(alice, spot2), at(bob, spot2), lost\_fight(alice) & 0 \\
			\hline 12 & at(alice, spot2), at(bob, spot1) & 0.11& 26 & at(alice, spot1), at(bob, spot1), lost\_fight(bob) & 0.01 \\
			\hline 21 & at(alice, spot2), at(bob, spot1) & 0.01 & 27 & at(alice, spot1), at(bob, spot1), lost\_fight(alice) & 0.01 \\
			\hline 22 & at(alice, spot1), at(bob, spot2) & 0.01 &&&\\
			\hline
		\end{tabular}
		\caption{}
		\label{tab:fishers-default-state-fluents}
	\end{subfigure}
	\caption{Semantics for the fishers default action situation: game tree (top) and state fluents with the probabilities induced over the terminal nodes (bottom).}
	\label{fig:fishers-default-semantics}
\end{figure}

\begin{listing}
\begin{minted}{prolog}
% the looser of the race cannot stay <==> must leave
rule(fishers,choice,1,if role(A,fisher) then ~can(A,stay)
	where [at(A,S),lost_race(A),fishing_spot(S)]).

% the winner of the race has to stay <==> cannot leave
rule(fishers,choice,1,if role(A,fisher) then ~can(A,leave)
	where [at(A,S),\+lost_race(A),fishing_spot(S)]).

rule(fishers,control,1,if does(F1,go_to_spot(S)) and does(F2,go_to_spot(S))
	then [lost_race(F1) withProb P1,lost_race(F2) withProb P2]
	where [F1@<F2,fishing_spot(S),speed(F1,X1),speed(F2,X2),{P1=X1/(X1+X2)},{P2=X2/(X1+X2)}]).
\end{minted}
\caption{Additional rules with priority 1 for the {\em first-in-time, first-in-right} rule configuration of Ostrom's fishing game.}
\label{code:fishing-game-description-race}
\end{listing}

\begin{figure}[h]
	\centering
	\begin{subfigure}[h]{\textwidth}
		\centering
		\includegraphics[width=\textwidth]{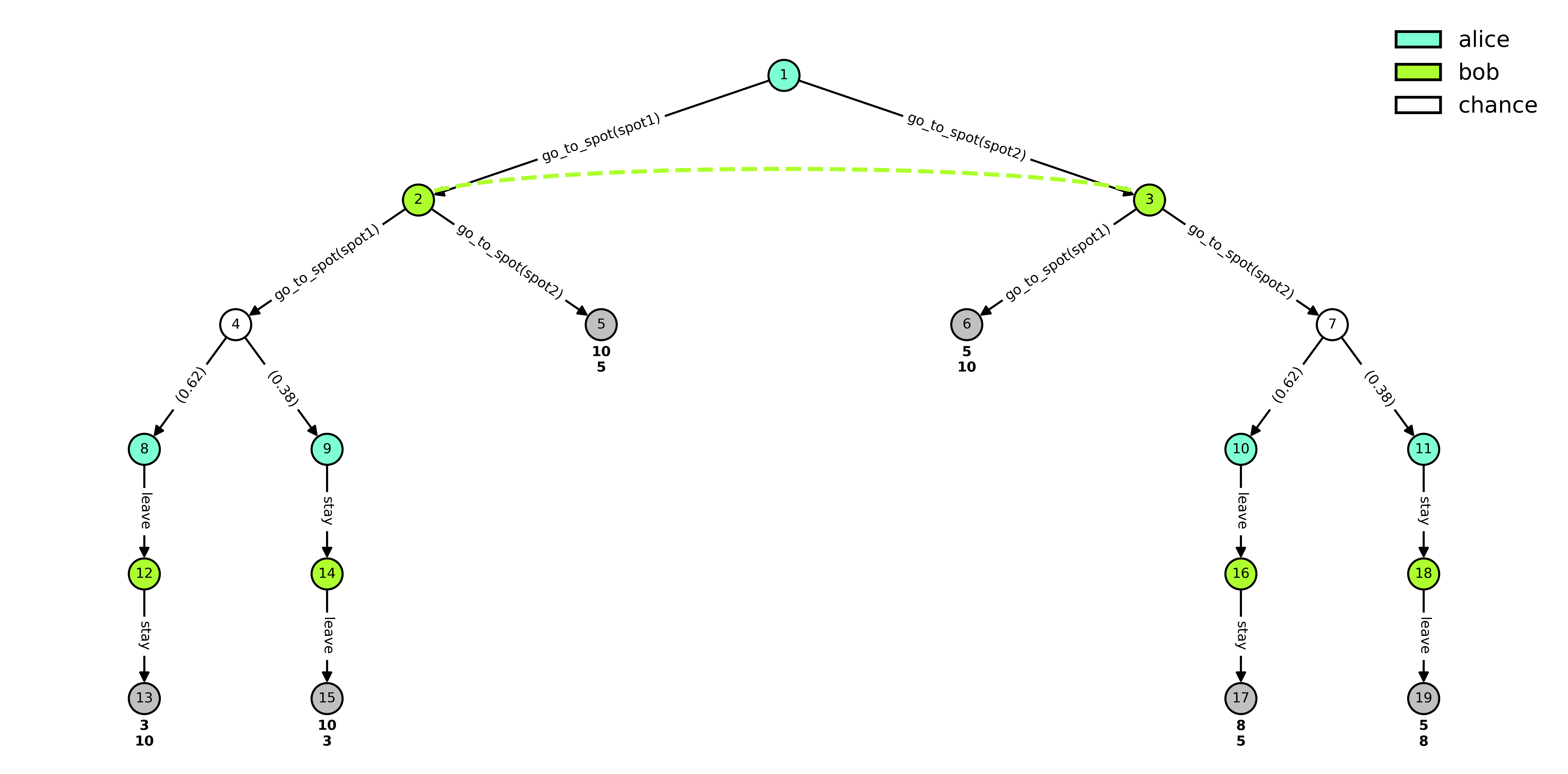}
		\caption{}
		\label{subfig:fishers-race-game-tree}
	\end{subfigure}
	
	\begin{subfigure}[h]{\textwidth}
		\centering
		\small
		\begin{tabular}{|c|l|c|}
			\hline & \textbf{State fluents} & $\mathbf{p}$\\
			\hline 1 & at(alice, shore), at(bob, shore) & - \\
			\hline 5 & at(alice, spot1), at(bob, spot2) & 0 \\
			\hline 6 & at(alice, spot2), at(bob, spot1) & 0 \\
			\hline 8 & at(alice, spot1), at(bob, spot1), lost\_race(bob) & - \\
			\hline 9 & at(alice, spot1), at(bob, spot1), lost\_race(alice) & - \\
			\hline 10 & at(alice, spot2), at(bob, spot2), lost\_race(bob) & - \\
			\hline 11 & at(alice, spot2), at(bob, spot2), lost\_race(alice) & - \\
			\hline 13 & at(alice, spot1), at(bob, spot2), lost\_race(bob) & 0.62 \\
			\hline 15 & at(alice, spot2), at(bob, spot1), lost\_race(alice) & 0.38 \\
			\hline 17 & at(alice, spot2), at(bob, spot1), lost\_race(bob) & 0 \\
			\hline 19 & at(alice, spot1), at(bob, spot2), lost\_race(alice) & 0 \\
			\hline
		\end{tabular}
		\caption{}
		\label{tab:fishers-race-state-fluents}
	\end{subfigure}
	\caption{Semantics for the fishers {\em first-in-time, first-in-right} action situation: game tree (top) and state fluents with the probabilities induced over the terminal nodes (bottom).}
	\label{fig:fishers-race-semantics}
\end{figure}

\begin{listing}
\begin{minted}{prolog}
% pick a random participant as the announcer
rule(fishers,position,2,if participates(A) then role(A,announcer)
	where [findall(X,participates(X),L),random_member(A,L)]).

% when both fishers at shore, the announcer makes one (and only one) announcement
rule(fishers,choice,2,if role(A,announcer) then can(A,announce_spot(S))
	where [at(A,shore),at(B,shore),A\=B,fishing_spot(S)]).
rule(fishers,choice,2,if role(A,announcer) then ~can(A,announce_spot(S))
	where [announced(A,_),fishing_spot(S)]).

% no-one can go for a spot before the announcement is made
rule(fishers,choice,2,if role(A,fisher) then ~can(A,go_to_spot(S))
	where [\+announced(_,_),fishing_spot(S)]).

% announcement is made
rule(fishers,control,2,if does(F,announce_spot(S)) then [announced(F,S) withProb 1] where []).

% if the fishers both go to the fishing spot declared by the announcer, the announcer is
% guaranteed to get there first
rule(fishers,control,2,if does(F1,go_to_spot(S)) and does(F2,go_to_spot(S))
	then [lost_race(F2) withProb 1] where [announced(F1,S),F1\=F2]).
\end{minted}
\caption{Additional rules with priority 2 for the {\em first-to-announce, first-in-right} rule configuration of Ostrom's fishing game.}
\label{code:fishing-game-description-announce}
\end{listing}

\begin{figure}[h]
	\centering
	\begin{subfigure}[h]{\textwidth}
		\centering
		\includegraphics[width=\textwidth]{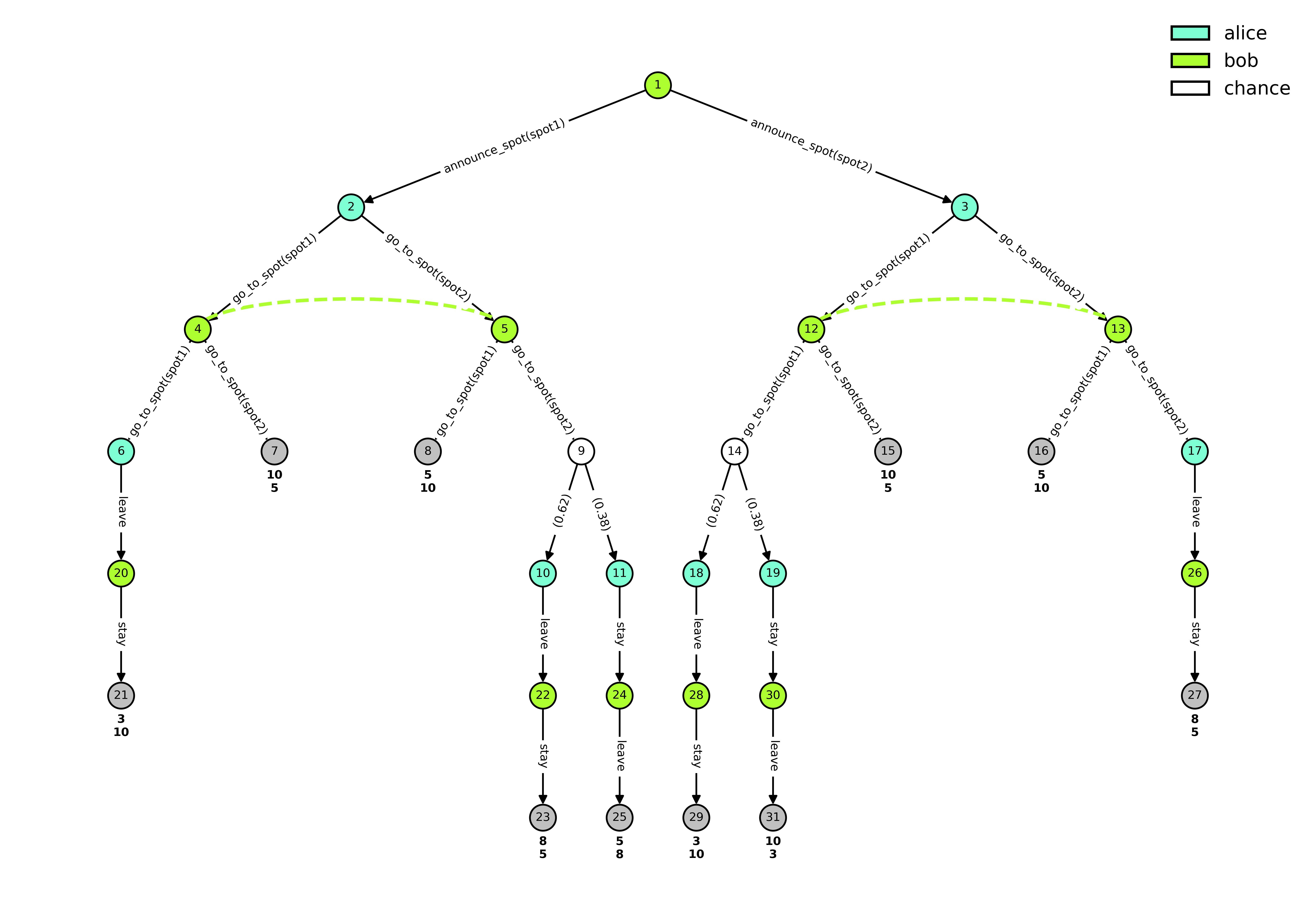}
		\caption{}
		\label{subfig:fishers-announce-game-tree}
	\end{subfigure}
	
	\begin{subfigure}[h]{\textwidth}
		\centering
		\small
		\begin{tabular}{|c|l|c|}
			\hline & \textbf{State fluents} & $\mathbf{p}$ \\
			\hline 1 & at(alice, shore), at(bob, shore) & - \\
			\hline 2 & announced(bob, spot1), at(alice, shore), at(bob, shore) & - \\
			\hline 3 & announced(alice, spot2), at(alice, shore), at(bob, shore) & - \\
			\hline 6 & announced(alice, spot1), at(alice, spot1), at(bob, spot1), lost\_race(bob) & - \\
			\hline 7 & announced(alice, spot1), at(alice, spot1), at(bob, spot2) & 0 \\
			\hline 8 & announced(alice, spot1), at(alice, spot2), at(bob, spot1) & 1.00 \\
			\hline 10 & announced(alice, spot1), at(alice, spot2), at(bob, spot2), lost\_race(bob) & - \\
			\hline 11 & announced(alice, spot1), at(alice, spot2), at(bob, spot2), lost\_race(alice) & - \\
			\hline 15 & announced(alice, spot2), at(alice, spot1), at(bob, spot2) & 0 \\
			\hline 16 & announced(alice, spot2), at(alice, spot2), at(bob, spot1) & 0 \\
			\hline 17 & announced(alice, spot2), at(alice, spot2), at(bob, spot2), lost\_race(bob) & - \\
			\hline 18 & announced(alice, spot2), at(alice, spot1), at(bob, spot1), lost\_race(bob) & - \\
			\hline 19 & announced(alice, spot2), at(alice, spot1), at(bob, spot1), lost\_race(alice) & - \\
			\hline 21 & announced(alice, spot1), at(alice, spot1), at(bob, spot2), lost\_race(bob) & 0 \\
			\hline 23 & announced(alice, spot1), at(alice, spot2), at(bob, spot1), lost\_race(bob) & 0 \\
			\hline 25 & announced(alice, spot1), at(alice, spot1), at(bob, spot2), lost\_race(alice) & 0 \\
			\hline 27 & announced(alice, spot2), at(alice, spot2), at(bob, spot1), lost\_race(bob) & 0 \\
			\hline 29 & announced(alice, spot2), at(alice, spot1), at(bob, spot2), lost\_race(bob) & 0 \\
			\hline 31 & announced(alice, spot2), at(alice, spot2), at(bob, spot1), lost\_race(alice) & 0 \\
			\hline
		\end{tabular}
		\caption{}
		\label{tab:fishers-announce-state-fluents}
	\end{subfigure}
	\caption{Semantics for the fishers {\em first-to-announce, first-in-right} action situation: game tree (top) and state fluents with the probabilities induced over the terminal nodes (bottom).}
	\label{fig:fishers-announce-semantics}
\end{figure}

\end{document}